\pdfoutput=1
\documentclass[11pt]{article}

\usepackage[]{ACL2023}

\usepackage{times}
\usepackage{latexsym}
\usepackage{graphicx}
\usepackage{multirow}
\usepackage{booktabs}
\usepackage[T1]{fontenc}
\usepackage[utf8]{inputenc}

\usepackage{microtype}

\usepackage{inconsolata}

\usepackage{amsmath, amssymb, amsfonts, amsthm}
\usepackage{bbm}
\usepackage{graphicx}
\usepackage{color}
\usepackage{bm}
\usepackage{float}
\usepackage{graphicx}
\usepackage{mathabx}
\usepackage{multirow}
\usepackage{setspace}
\usepackage{hyperref}
     
\usepackage{times}
\usepackage{latexsym}

\usepackage{url}
\usepackage{adjustbox}  
\usepackage{color, colortbl}
\usepackage{xspace}
\usepackage{tikz}
\usetikzlibrary{arrows,shapes,automata,backgrounds,petri}

\newcommand*{\ie}{i.e.\@\xspace}

\newcommand\ModelName{Z-Code++}

\newcommand{\modelp}[2]{#1{$_{\mathtt{#2}}$}}
\newcommand{\mlm}{\mathtt{MLM}}
\newcommand{\rtd}{\mathtt{RTD}}
\newcommand{\csp}{\mathtt{CSP}}

\newcommand{\zcodepp}{Z-Code++}
\newcommand{\zcodepplarge}{\zcodepp$_{\mathtt{LARGE}}$}

\title{ \ModelName: A Pre-trained Language Model Optimized for Abstractive Summarization}

\author{Pengcheng He$^1$, Baolin Peng$^2$, Song Wang$^1$, \\
\bf{Yang Liu$^1$, Ruochen Xu$^1$, Hany Hassan Awadalla$^1$,Yu Shi$^1$, Chenguang Zhu$^1$,}\\
 \bf{Wayne Xiong$^1$, Michael Zeng$^1$, Jianfeng Gao$^2$, Xuedong Huang$^1$} \\
    $^1$ Microsoft Azure AI \\
    $^2$ Microsoft Research \\
 {\tt penhe@microsoft.com}
}

\date{November 2021}
\graphicspath{figures/}
\DeclareGraphicsExtensions{.png,.pdf,.svg}

\begin{document}

\maketitle

\begin{abstract}

This paper presents Z-Code++, a new pre-trained language model optimized for abstractive text summarization. 
The model extends the state of the art encoder-decoder model using three techniques. First, we use a two-phase pre-training process to improve model's performance on low-resource summarization tasks. The model is first pre-trained using text corpora for language understanding, and then is continually pre-trained on summarization corpora for grounded text generation. Second, we replace self-attention layers in the encoder with disentangled attention layers, where each word is represented using two vectors that encode its content and position, respectively. Third, we use fusion-in-encoder, a simple yet effective method of encoding long sequences in a hierarchical manner.
Z-Code++ creates new state of the art on 9 out of 13 text summarization tasks across 5 languages. Our model is parameter-efficient in that it outperforms the 600x larger PaLM\textsubscript{540B} on XSum, and the finetuned 200x larger GPT3\textsubscript{175B} on SAMSum. In zero-shot and few-shot settings, our model substantially outperforms the competing models. %Z-Code++ models also perform well on a broad range of natural language understanding and generation tasks.

\end{abstract}

\section{Introduction}

Text summarization aims at producing a concise and fluent summary while preserving salient content and overall meaning of the source documents. It has been applied in a wide range of real-world applications, e.g., summarizing Web search results for interactive information retrieval \citep{gao2022neural} and generating medical summaries from doctor-patient conversation transcripts \citep{zhang2021leveraging}.

While the \emph{extractive} approach is the dominant approach in commercial systems due to its simplicity and effectiveness \citep{allahyari2017text}, the \emph{abstractive} approach is getting more attention in the research community as neural language models are used \citep[e.g.,][]{rush2015neural, nallapati2016abstractive, chopra2016abstractive, liu2019text, liu2019hierarchical, pasunuru2021data}. 
Compared to the extractive approach where a summary is constructed using extracted sentences, abstractive summarizers paraphrase the idea of the source documents in a new form, and have a potential of generating more concise and coherent summaries. 

However, good abstractive summarizers are harder to develop since we have to deal with problems like semantic representation, inference and low-resource text generation, which are more challenging than sentence extraction. 
Recently, large-scale pre-trained language models (PLMs) such as PEGASUS \citep{zhang2020pegasus}, GPT \citep{gpt2,gpt3}, T5 \citep{raffel2019t5}, have been applied for abstractive summarization. While these models can produce surprisingly fluent text, the generated summaries often contain factual inconsistencies, caused by distorted or fabricated facts about the source documents, which is known as the \emph{hallucination} problem \citep{kryscinski2019neural,celikyilmaz2020evaluation,ji2022survey}. 
In addition, since the amount of text in the source documents can be very large, it is expensive to train an end-to-end abstractive model (e.g., an encoder-decoder transformer model) given the memory constraints of current hardware and the latency constraints of applications such as online document summarization for interactive information retrieval. Therefore, a two-stage approach is widely used, where a subset of document sentences is coarsely selected using an extractive summarizer, and an abstractive summarizer generates the summary conditioning on the extraction \citep{liu2019text}.  This approach is sub-optimal in that salient information might be missed in the extraction.

In this paper, we propose a new encoder-decoder PLM optimized for abstractive summarization, {\ModelName}, which significantly extends Z-Code~\citep{wang2020multi}, a state-of-the-art PLM developed for machine translation, as follows.  

First, {\ModelName} is pre-trained on web text using two tasks, replaced token detection (RTD) and corrupted span prediction (CSP). 
RTD uses a generator to generate ambiguous corruptions and a discriminator to distinguish the ambiguous tokens from the original inputs \citep{clark2020electra}.  RTD is proved to be more sample-efficient than the classic mask language modeling (MLM) task in learning text representations for language understanding \citep{bajaj2022metro, hao2021learning}. 
In CSP, a consecutive segment of tokens are corrupted and the model is learned to predict the corrupted spans using all the uncorrupted tokens in the original input \citep{raffel2019t5,joshi2019spanbert}. CSP can be viewed as a generalized form of gap sentences generation (GSG), a pre-training task tailored to abstractive summarization \citep{zhang2020pegasus}, where the spans are entire sentences. CSP outperforms GSG in our experiments. In the second phase of grounded pre-training~\citep{GODEL}, the model is continually trained on summarization corpora of documents-summary pairs to better support low-resource fine-tuning to downstream summarization tasks that require the model to produce summaries grounded in source documents. We find in our experiments that grounded pre-training significantly boosts the results on downstream tasks in low-resource settings.

To handle the large input documents, we use fusion-in-encoder (FiE), a simple yet effective method of encoding long sequences in a hierarchical manner. It works by first splitting the input sequence into small chunks, applying attention on each chunk locally to get the chunk representation, and applying attention globally on the concatenated chunk representations to get the representation of the original input. 

In addition, we replace the self-attention layer in the encoder with the disentangled attention (DA) layer~\citep{he2020deberta,he2021debertav3}, where each word is represented using two vectors that encode its content and position, respectively, and the attention weights among words are computed using disentangled matrices on their contents and relative positions, respectively.  DA is motivated by the observation that the attention weight of a word pair depends on not only their contents but their relative positions. For example, the dependency between the words ``deep'' and ``learning'' is much stronger when they occur next to each other than when they occur in different sentences. We show in our experiments that DA leads to a more effective abstractive summarizer.

For evaluation, we have pre-trained two {\ModelName} models on English data and multi-lingual data, respectively. The English model is trained using 160G English text data and the vocabulary of DeBERTaV2 ~\citep{he2020deberta}. The multi-lingual model is trained on mC4 corpus which is the same as mT5. These models are evaluated on 13 text summarization tasks across 5 languages, and create new state of the art on 9 tasks. 
As of May 6th, 2022, {\ModelName} sits atop of the XSum leaderboard, surpassing UL2\textsubscript{20B}, T5\textsubscript{11B} and PEGASUS.
It is worth noting that our models are very parameter-efficient. For example,  
{\ModelName} outperforms PaLM\textsubscript{540B}, which is 600x larger in model parameters, on XSum, and outperforms a fine-tuned, 200x larger, GPT3\textsubscript{175B} on SAMSum.
In zero-shot and few-shot settings, our models outperform more substantially the competing models.

%We also evaluate the {\ModelName} models on other 13 natural language understanding and generation tasks across 20 languages, including WebNLG,E2E NLG, XNLI, TyDiQA, SQuAD, and 8 GLUE tasks of text classification. 
%Our models create new state of the art on WebNLG and E2E NLG, and compare favorably with other state-of-the-art PLMs on the GLUE benchmark.
%On the multilingual tasks such as XNLI and TyDiQA, m{\ModelName} outperforms mT5\textsubscript{large} and XLM-R\textsubscript{large}.
%Similar to the evaluation results on summarization tasks, {\ModelName} outperforms the 600x larger PaLM\textsubscript{540B} on WebNLG-En and E2E NLG, demonstrating its parameter-efficiency. 

%\input{1_introduction.tex}

\section{Z-Code++}
\label{sec:method}

This section describes three modeling techniques we have exploited to optimize Z-Code++ for abstractive summarization, including two-phase pre-training, disentangled attention, and long sequence encoding.

\subsection{Two-Phase Pre-Training}
\label{subsec:two-phrase-pre-training}

The two-phase pre-training, which includes the \emph{language model pre-training} and \emph{grounded pre-training} phases, is inspired by the GODEL recipe \citep{GODEL} that has been proposed to pre-train language models for grounded text generation tasks, such as dialog response generation and abstractive question-answering. 

\begin{figure*}[htb!]
\centering  
\includegraphics[width=0.93\linewidth]{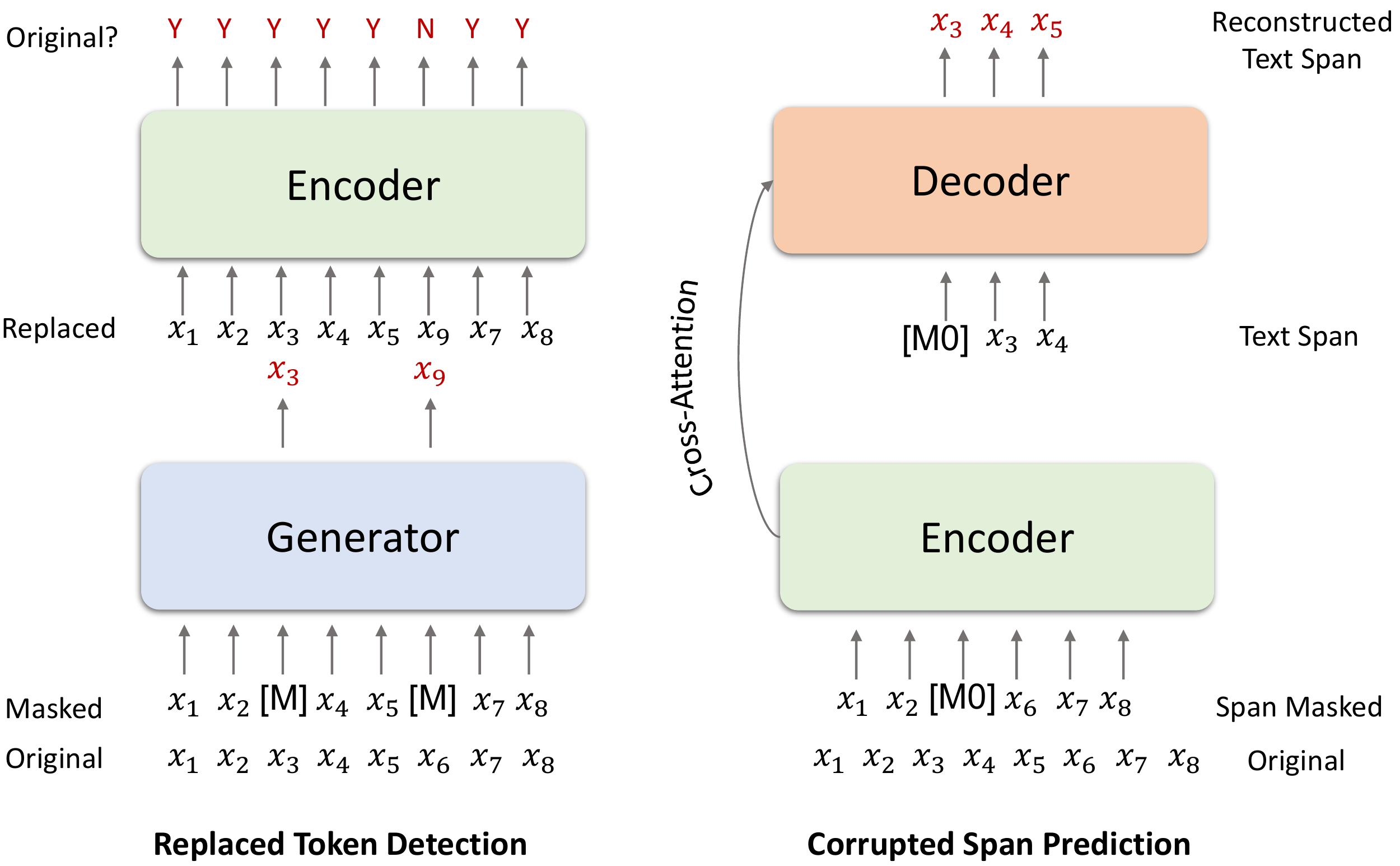}
\caption{The two pre-training tasks, replaced token detection (RTD) and corrupted span prediction (CSP), used in the language model pre-training phase of {\ModelName}. RTD task is to optimize the encoder, and CSP is to optimize the encoder-decoder. Encoders in the same color share parameters during training.}
\label{fig:zpp}
\end{figure*}

In the \emph{language model pre-training} phase, Z-Code++ is pre-trained using two language modeling tasks, replaced token detection (RTD) \citep{clark2020electra} and corrupted span prediction (CSP) \citep{raffel2019t5,joshi2019spanbert}. 
As illustrated in Figure\ref{fig:zpp} (Left), RTD uses a generator trained with MLM to generate ambiguous tokens to replace tokens in the original input $\bm{X}$, and a discriminator to determine whether a token is from $\bm{X}$ or generated by the generator. 
Let $\theta _ G$ and $\theta _ D$ be the parameters of the generator and the discriminator, respectively. The MLM loss of the generator is written as  

\begin{equation}
\resizebox{.85\hsize}{!}{$%
\label{eq:lg}
    L_{\mlm} =  \mathbbm{E}\left( -\sum_{i\in \mathcal{C}}{\text{log} \;  p_{\theta_G}\left(\tilde{x}_{i,G}=x_i|\bm{\tilde{X}}_G\right)} \right),
    $}
\end{equation}
where $\bm{\tilde{X}}_G$ is the input to the generator by randomly masking $15\%$ tokens in original input $\bm{X}$. The input sequence of the discriminator is constructed by replacing the masked tokens, $x_i, i \in \mathcal{C}$, with the tokens, $\tilde{x}_{i}$, sampled by the generator as
\begin{equation}
\label{eq:xd}
\resizebox{.85\hsize}{!}{$
\tilde{x}_{i,D} = \left\{ \begin{array}{cr}
    \tilde{x}_{i} \sim  p_{\theta_G}\left(\tilde{x}_{i,G}=x_i|\bm{\tilde{X}}_G\right),  & i \in \mathcal{C}  \\
      x_{i}, & i \notin \mathcal{C}.
\end{array}\right.
$}
\end{equation}

Then the discriminator is trained using the loss
\begin{equation}
\resizebox{.85\hsize}{!}{$
\label{eq:ld}
    L_{\rtd} =    \mathbbm{E}\left(- \sum_{i}{\text{log} \;  p_{\theta_D}\left(\mathbbm{1}\left(\tilde{x}_{i,D}=x_i\right)|\bm{\tilde{X}}_{D}, i \right)}\right),
$}
\end{equation}
where $\mathbbm{1}(\cdot)$ is the indicator function and $\bm{\tilde{X}}_{D}$ is the input to the discriminator constructed via \eqref{eq:xd}. 
In ELECTRA~\citep{clark2020electra}, the discriminator and generator share token embeddings and their parameters are optimized via MLM and RTD jointly as $L = L_{\mlm} + \lambda L_{\rtd}$. 
However, as pointed out in \cite{he2021debertav3}, such embedding sharing makes training highly inefficient since MLM and RTD pull token embeddings into very different directions, creating the ``tug-of-war'' dynamics. MLM tries to map the tokens that are semantically similar to the embedding vectors that are close to each other. RTD, on the other hand, tries to discriminate semantically similar tokens, pulling their embeddings as far as possible to optimize the classification accuracy. Thus, we use the method of gradient-disentangled embedding sharing~\citep{he2021debertav3} by re-parameterizing the token embeddings of the discriminator as
\begin{equation}
\label{eq:de}
    \bm{E}_D = sg(\bm{E}_G) + \bm{E}_\Delta,
\end{equation}
where $\bm{E}_D$ and $\bm{E}_G$ are the embedding parameters of the discriminator and generator, respectively, $sg$ is the stop gradient operator which only allows gradients propagation through $\bm{E}_\Delta$.
$\bm{E}_\Delta$ is initialized as a zero matrix.   
In each training pass, we first run a forward pass of the generator to generate inputs for the discriminator, and then a backward pass to update $\bm{E} _ G$ with respect to MLM.
After that, we run a forward pass for the discriminator using the inputs produced by the generator and run a backward pass with respect to the RTD loss to update $\bm{E} _ D$ by propagating gradients only through $\bm{E}_\Delta$. After model training, $\bm{E}_\Delta$ is added to $\bm{E}_G$ and the sum is saved as $\bm{E}_D$ in the discriminator, as Equation~\ref{eq:de}.

%Z-Code++ adopts RTD to improve the efficiency of encoder-decoder model with the encoder learns a better representation of the context.

The CSP is widely used to optimize the encoder-decoder PLMs such as T5~\citep{raffel2019t5}. As illustrated in Figure~\ref{fig:zpp} (Right), given input string $\bm{X}$, we first select a continuous span $\bm{Y}_i$ by first randomly selecting a start position in $\bm{X}$ and a span with an average length of 3. Then we replace the selected span $\bm{Y}_i$ with a sentinel token \texttt{[M$_i$]}. We repeat the process until the replaced tokens amount to 15\% of all tokens in $\bm{X}$. Then, we feed the corrupted input $\bm{\tilde{X}}_{CSR}$ to the encoder. The encoder-decoder model is then trained to recover the $\bm{Y}_i$ from the context. The CSP loss is written as
\begin{equation}
\label{eq:csp}
    L_{\csp} =  \mathbbm{E}\left( -\sum_{i=1}^{|\bm{Y}|}{\text{log} \;  p_{\theta}\left(\bm{Y}_i|\bm{\tilde{X}}_{\csp}, \bm{Y}_<i\right)} \right)
\end{equation}
If we restrict the corrupted span $\bm{Y}_i$ to a complete sentence, CSP is equivalent to the GSG task which simulates the process of extractive summarization and is shown to be effective for training abstractive summarizers~\citep{zhang2020pegasus}. In this study, we find the that CSP, as a more general form of GSG, works better across many natural language understanding and generation tasks, including summarization, as to be discussed in Section~\ref{sec:experiment}.

Combining the pre-training tasks of MLM, RTD and CSP, in the language model pre-training phase, {\ModelName} is optimized using the joint loss as 
$L = \lambda_{1} L_{\mlm} + \lambda_{2} L_{\rtd} + \lambda_{3} L_{\csp}$, where we set $\lambda_{1}=1, \lambda_{2}=30, \lambda_{3}=1$ in our experiment.

In the second phase of \emph{grounded pre-training}, {\ModelName} is continually pre-trained on a collection of summarization datasets, as shown in Table~\ref{tab:grounded_pretraining}, which consist of documents-summary pairs $(\bm{X}, \bm{Y})$, to better support low-resource finetuning for downstream summarization tasks that require the model to generate target summaries $\bm{Y}$ grounded in source documents $\bm{X}$, as
\begin{equation}
p(\bm{Y}|\bm{X}) = \prod_{n=1}^N p(y_n|y_1, \cdots, y_{n-1}, \bm{X})
\label{eq:lm}
\end{equation}
Following T0~\citep{wei2021finetuned}, FLAN~\citep{sanh2022multitask}, and GODEL~\citep{GODEL}, we add for each training pair $(\bm{X}, \bm{Y})$ a natural language instruction of the summarization task, as illustrated in the below example and in Table~\ref{tab:grounded_pretraining}.  In our experiment, we only apply \emph{grounded pre-training} for low-resource summarizations. Unless specified, we apply the first phase \zcodepp{} to downstream task adaptation.

% refers to language model pre-training phase model. 

\begin{figure}[htb!]
\centering  
\includegraphics[width=1.0\linewidth]{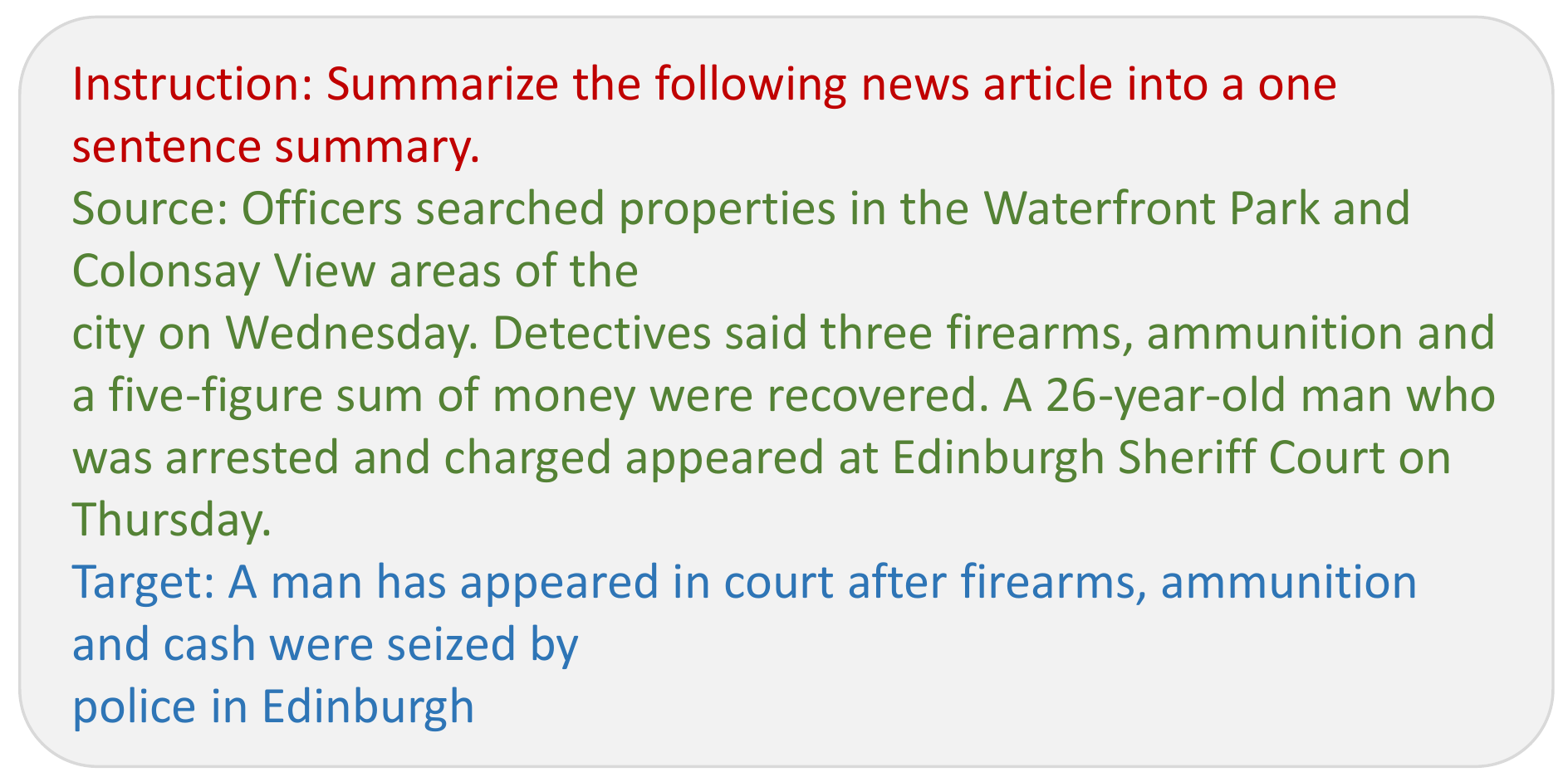}
% \caption{Examples of instructions and grounded pre-training datasets.}
\caption{Examples of instructions used for grounded pre-training.}
\label{fig:example}
\end{figure}

\begin{table*}[htb!]
    \centering
    %\begin{adjustbox}{max width=0.95\textwidth}
    \scalebox{0.7}{
    \begin{tabular}{@{\hskip1pt}l| c| l @{\hskip1pt}}
        \toprule
        Task & Genre & Instructions \\
        \midrule
        MediaSum & Interview & Summarize the following interview script into a two sentences summary. \\
            & - &  How can the following interview script be rephrased into a few sentences summary. \\ \hline
        MultiNews & News & Summarize the news article into a one sentence summary.  \\
        & - & Rephrase the news article with a few sentences.\\ \hline
        NewsRoom & News & Summarize the news article into a one sentence summary.  \\
        & - & Rephrase the news article concisely with a few sentences.\\ \hline
        WikiHow & Wiki & Summarize the paragraph into a one sentence summary. \\
        & - & Summarize the paragraph with a few words.\\ 
        \bottomrule
        \end{tabular}}
    %\end{adjustbox}
    \caption{
    Grounded pre-training summarization datasets and examples of instructions.
    }
    \label{tab:grounded_pretraining}
    %\vspace{-3mm}
\end{table*}

\subsection{Disentangled Attention}
\label{subsec:disentangled-attention}

Disentangled Attention (DA) is first used in DeBERTa ~\citep{he2020deberta,he2021debertav3}.
DA is an extension of the classic self-attention (SA) mechanism in that DA represents each input word using two separate vectors: one for the content and the other for the position. Meanwhile, its attention weights among words are computed via disentangled matrices on both their contents and relative positions. The experiments of DeBERTa shows that DA is more efficient than SA to encode the positional dependency in Transformer models. {\ModelName} adopts DA in modeling. Our experiments show that DA leads to a more effective abstractive summarizer.

\subsection{Long Sequence Encoding}
\label{subsec:long-sequence-encoding}
It is challenging to encode long sequence given the $O(N^2)$ memory and computation complexity of self-attention and DA. Various sparse attention mechanisms have been proposed to alleviate the problem. However, sparse attention often hurts performance on short sequences due to the decrease of attention precision. Inspired by fusion-in-decoder~\citep{izacard2020leveraging} and hierarchical transformer~\citep{liu2019hierarchical}, we propose fusion-in-encoder (FiE), a simple but effective mechanism to encode long sequences while retaining high attention precision on short sequences. FiE works by separating the $L$ encoder layers of {\ModelName} into $m$ local layers and $n$ global layers. In each local layer, the hidden states of input sequence are split into small chunk of size $l$ (e.g. 256 or 512), and self-attention (or DA) is only applied to those small chunks locally with a complexity of $O(l^2)$. After local layer, the hidden states of those small chunks are concatenated together to form the representation of the long sequence. Global layers are the same as original self-attention (or DA) layers in encoder to fuse the local states of small chunks. With FiE, the complexity of encoder is reduced from $O(LN^2)$ to $O(mNl+nN^2)$. Both the local layers and fusion layers are initialized with the corresponding weights of encoder layers of {\ModelName}. Please check Appendix \ref{app:fie} for a graphic illustration of FiE.
%If we set $l$ to be larger than the input sequence length, then FIE is reduced to the same as original encoder and the performance keeps the same. 
In experiment, we show that compared with LongT5~\citep{guo2021longt5} which applies sparse attention that is specifically optimized for summarization, {\ModelName} achieves similar or better performance on long document summarization tasks.

\section{Experiment}
\label{sec:experiment}

\subsection{Experiment Setups}
\label{sec:setup}
\begin{table*}[htb!]
\centering
\scalebox{0.7}{
\begin{tabular}{lcccc}
\toprule
% Dataset     & \# Docs. & \# Input Tokens & \# Summary Tokens & Genre              \\
            % &       &Avg/95\%                   &   Avg/95\%& \\
\multirow{2}{*}{\bf{Dataset}} &          \multirow{2}{*}{\# Docs. }  & \# Input Tokens & \# Summary Tokens & \multirow{2}{*}{Genre} \\
&       &Avg/95\%                   &   Avg/95\%& \\
\midrule
\multicolumn{5}{l}{\emph{Standard Document Summarization}} \\
\midrule
AESLC       & 14K      & 152/440             & 5/13                & Business/Personal \\
SAMSum      & 15k     & 132/331             & 24/52               & Dialog             \\
XSUM        & 227K     & 458/1,139            & 25/35               & News           \\
WikiHow     & 168K     & 623/1,878            & 90/226               & Wiki              \\
NewsRoom    & 1.3M     & 715/1,704            & 43/152               & News              \\
CNNDM       & 311K     & 827/1,682            & 74/127               & News              \\
Reddit TIFU & 41K     &  470/1,096           &  24/51              & Forum             \\
\midrule
\multicolumn{5}{l}{\emph{Long Document Summarization}} \\ 
\midrule
MediaSum    & 463K     & 1,554/5,323          & 14/52               & Interview         \\
MultiNews  & 459K     & 2,103/6,642           & 264/407              & News              \\
PubMed      & 133K     & 3,224/8,210          & 214/401              & Scientific       \\
arXiv       & 215K     & 6,913/19,560          & 293/576              & Scientific        \\
\midrule
\multicolumn{5}{l}{\emph{Multilingual Summarization}} \\ 
\midrule
WikiLingua (ru $\rightarrow$ en)  & 37K      & 661/1,468            & 49/102               & Wiki              \\
WikiLingua (vi $\rightarrow$ en)  & 13K      & 1,140/2,570            & 48/96               & Wiki              \\
WikiLingua (es $\rightarrow$ en)  & 79K      & 676/1,454            & 50/105               & Wiki              \\
WikiLingua (tr $\rightarrow$ en)  & 3k      & 549/1,294            & 50/100               & Wiki              \\
MLSum (de)       & 221k     & 907/1,712       & 50/81            & News                 \\
MLSum (es)       & 266k     & 1,195/2,402      & 31/50            & News                  \\
\bottomrule
\end{tabular}}
\caption{Statistics of the datasets used for evaluation including the total number of documents, the average length of input tokens and summary tokens, and the genres of each dataset.}
\label{table:datasets}
\end{table*}
\paragraph{Datasets} We validate the effectiveness of \ModelName{} on 11 representative summarization tasks, which are detailed in Table \ref{table:datasets}.
Among these datasets, XSum~\citep{narayan2018don}, CNNDM~\citep{see2017get}, NewsRoom~\citep{grusky2018newsroom}, and MultiNews~\citep{fabbri2019multi} are news article summarizations, while SAMSum~\citep{gliwa2019samsum}, MediaSum~\citep{zhu2021mediasum}, and Reddit TIFU~\citep{kim2018abstractive} are conversation-like summarization tasks. Following LongT5, we use MultiNews, MediaSum, arXiv~\citep{cohan2018discourse} and PubMed~\citep{cohan2018discourse}  to assess the long document summarization capability. In addition, WikiLingua~\citep{ladhak2020wikilingua} and MLsum~\citep{scialom2020mlsum} are used to evaluate the capacity of \ModelName{} on multilingual summarization.
\paragraph{Implementation Details} We have built our models following the same setting as T5. For \zcodepplarge{}, there are 24 layers for the encoder and 24 layers for the decoder with 1024 hidden dimension sizes and 16 self-attention heads. Following DeBERTaV3~\cite{he2021debertav3}, a 6-layer generator with the same structure as the encoder is employed during the pre-training stage. \zcodepplarge{} is trained on 160G data with a vocabulary of size 128k. Our code is implemented based on open sourced pytorch\footnote{\url{https://pytorch.org/}} and DeBERTa\footnote{\url{https://github.com/microsoft/DeBERTa}}. We pre-train \zcodepplarge{} for 1M steps with a batch size of 2048 in Azure Machine Learning cluster\footnote{\url{https://ml.azure.com}} with 128 A-100 GPUS for 20 days. AdamW is used as the optimizer in all experiments. For tasks with an input length of more than 10k words, i.e., arXiv and PubMed, Fusion-in-Encoder is used to encode the document as described in \ref{subsec:long-sequence-encoding}. For the other standard summarization tasks with moderate input length (i.e., less than 4k words) we directly feed the input document to the encoder. 

% For multilingual summarization, we build m\zcodepplarge{} with the same model structure as \zcodepplarge{} except for the training data and vocabulary. Specifically, m\zcodepplarge{} is trained with mC4 data and a vocabulary size of 250k, which are the same as mT5\citep{xue2021mt5}. Following XLM\citep{lample2019xlm}, CCMatrix\citep{schwenk2019ccmatrix} and CCAligned\citep{el2019ccaligned} parallel data are used to enhance the cross-lingual transfer performance of m\zcodepplarge{}. Contrary to mT5, due to computation resource limitations, m\zcodepplarge{} is trained with only 500B tokens instead of 1T tokens. 

For multilingual summarization, we have built \zcodepplarge{} with the same architecture but different training data and vocabulary. Specifically, \zcodepplarge{} is trained with mC4 data and a vocabulary of size 250k, which are the same as mT5~\citep{xue2021mt5}. Following XLM~\citep{lample2019xlm}, CCMatrix~\citep{schwenk2019ccmatrix} and CCAligned~\citep{el2019ccaligned}, parallel data is used to enhance the cross-lingual summarization of \zcodepplarge{}. Due to the limited computational resource, \zcodepplarge{} is trained with only 500B tokens instead of 1T tokens as that for mT5 training. 

We use grid search to choose the grounded training and fine-tuning hyper-parameters based on validation set, the parameter search range are listed in appendix \ref{grid}. 

\subsection{Experiment Results}

\subsubsection{Results on Standard English Summarization Tasks}
 
\newcolumntype{g}{>{\columncolor{Gray}}c}
\definecolor{Gray}{gray}{0.97}
\begin{table}[htb!]
	\centering
	% \begin{adjustbox}{max width=0.99\columnwidth}
		\renewcommand{\arraystretch}{1.05}
  \scalebox{0.75}{
		\begin{tabular}{l|c|cg}
			\toprule
			% \multirow{2}{*}{\bf Dataset} & \multirow{2}{*}{Prior SOTA} & \multirow{2}{*}{ \shortstack[c]{\modelp{PEGASUS}{LARGE} &   \\ 470M }}  & \zcodepplarge{} \\ 
			\multirow{2}{*}{\bf Dataset} & \multirow{2}{*}{Prior SOTA}    & \modelp{PEGASUS}{LARGE} & \zcodepplarge{} \\ 
			                             &                                & 470M                    & 710M            \\ 
			\midrule
			XSum                         & 27.1\rlap{\textsuperscript{a}} & 24.6                    & \bf 24.6        \\      
			CNNDM                        & 22.6\rlap{\textsuperscript{b}} & 21.4                    & \bf 22.2 \footnotemark        \\      
			NewsRoom                     & 33.5                           & \bf 33.5                & 33.1            \\ 
			WikiHow                      & 18.5                           & 18.5                    & \bf22.1         \\   
			SAMSum                       & 29.8\rlap{\textsuperscript{c}} & 26.3                    & \bf30.3         \\
			Reddit TIFU                  & 11.3\rlap{\textsuperscript{d}} & 9.0                     & \bf11.6         \\  
			AESLC                        & 21.2                           & 21.2                    & \bf22.5         \\    
   \midrule
                \textbf{Average}                        & 23.4                           & 22.1                    & \bf23.8         \\    
			\bottomrule
		\end{tabular}
  }

	\caption{Results on Common English Summarization tasks. Best numbers are in \textbf{Bold}. \textsuperscript{a}ST-MOE\textsubscript{268B} \citep{zoph2022designing}, \textsuperscript{b}T5\textsubscript{11B} \citep{rothe2021thorough}, \textsuperscript{c}GPT3\textsubscript{175B+LoRA} \citep{hu2021lora},
	\textsuperscript{d}MAPPET+BART\textsubscript{LARGE} \citep{aghajanyan2021muppet}. 
  }
			
	\label{tab:en_sum}
	% \vspace{-2mm}
\end{table}

 We first conduct experiments to compare the performance of \zcodepplarge{} with SOTA and \modelp{PEGASUS}{LARGE} on 7 representative standard public English summarization datasets with moderate document length, including AESLC, SAMSum, XSUM, WikiHow, NewsRoom, CNN/DailyMail(CNNDM), and Reddit TIFU. Following \citep{chowdhery2022palm, gehrmann2022repairing}, for each dataset we report the average F-measure ROUGE-2 score of 5 runs. Detailed F-measure of ROUGE-1/ROUGE-2/ROUGE-L scores can be found in Appendix \ref{tab:mnlg}. 

 As listed in Table \ref{tab:en_sum} , \zcodepplarge{} achieves substantial improvements over \modelp{PEGASUS}{LARGE}, which is a PLM optimized for abstractive summarization,
%specifically designed and pre-trained with summarization-like objective, 
on 6 out of 7 tasks in terms of ROUGE-2 F-measure score. \footnote{The computation cost of the embedding layer is not factored in, so we only display the primary model parameters in the table, excluding those from the embedding layer. This approach is consistent across all subsequent experiments for comparison purposes.} Specifically, on SAMSum, a critical dialog summarization task, \zcodepplarge{} outperforms \modelp{GPT-3}{175B} that is extensively fine-tuned with LoRA\citep{hu2021lora} even though \zcodepplarge{} has less than 1/175 parameters of \modelp{GPT-3}{175B}. Furthermore, \zcodepplarge{} lifts SOTAs by 0.36 points on average. These results demonstrate the  effectiveness of \ModelName{} on English document summarization tasks. Additionally, we observe that \zcodepplarge{} outperforms \modelp{PEGASUS}{LARGE} on WikiHow, SAMSum, Reddit TIFU, and AESLC by a much larger margin ($>1\%$) than it does on XSum, CNNDM, and NewsRoom. We speculate that \modelp{PEGASUS}{} is biased to news-like tasks since it is heavily pre-trained on large amounts of news data. In contrast, \ModelName{} is pre-trained on diverse web data and thus is more adaptable for general-domain summarization tasks. \footnotetext{We have achieved 24.1 R2 score on CNNDM using exposure debiasing to address the mismatch between teacher forcing and student forcing learning, as we will describe in detail in a future publication.}

\subsubsection{Results on Long Document Summarization}
\newcolumntype{g}{>{\columncolor{Gray}}c}
\definecolor{Gray}{gray}{0.97}
\begin{table*}[htb!]
% \small
	\centering
	\renewcommand{\arraystretch}{1.0}
	% \begin{adjustbox}{max width=0.99\columnwidth}
		% \begin{tabular}{@{\hskip3pt}l@{\hskip2pt}|@{\hskip2pt} c@{\hskip2pt}||@{\hskip2pt} c@{\hskip2pt}|@{\hskip2pt} c@{\hskip2pt}|@{\hskip2pt} c@{\hskip2pt}| @{\hskip2pt}c@{\hskip2pt}}
  \scalebox{0.85}{
		\begin{tabular}{l|c|cccg}
						    
			% SOTA&LongT5\textsubscript{xlarge}&LongT5\textsubscript{large}&PEGASUS\textsubscript{large} & Z-Code++ \\
			\toprule
			\multirow{2}{*}{\bf Dataset} & \multirow{2}{*}{ Prior SOTA}   & \modelp{LongT5}{XLARGE} & \modelp{LongT5}{LARGE} & \modelp{PEGASUS}{LARGE} & \zcodepplarge{} \\
			                             &                                & 3B                      & 705M                     & 470M                    & 710M            \\ 
			%   \#Training Tokens& & & 500B& 1T& 1T& 500B \\
			\midrule
			MediaSum                     & 19.7                           & 19.7                    & 19.0                   & -                       & \bf20.2         \\  
			MultiNews                    & 21.1\rlap{\textsuperscript{a}} & 19.4                    & 18.4                   & 18.7                    & \bf21.6         \\ 
			arXiv                        & 21.9\rlap{\textsuperscript{b}} & 21.9                    & 20.6                   & 17.2                    & \bf22.5         \\    
			PubMed                       & 24.8                           & 24.8                    & 24.7                   & 19.6                    & \bf24.9         \\  
   \midrule
   	\textbf{Average}                       & 21.9                           & 21.5                    & 20.7                   & 18.5                    & \bf22.2         \\  
			\bottomrule
		\end{tabular}
  }
	% \end{adjustbox}
	\caption{
		Comparison results on long input summarization tasks. Best numbers are in \textbf{Bold}. \textsuperscript{a}PRIMER \citep{xiao2021primer}, \textsuperscript{b}Top-Down Transformer \citep{pang2022long}
	}
	\label{tab:en_long}
	% \vspace{-2mm}
\end{table*}
\begin{table*}[htb!]
	\centering
	%\begin{adjustbox}{max width=0.95\textwidth}
 \scalebox{0.88}{
		\begin{tabular}{@{\hskip3pt}l@{\hskip2pt}|@{\hskip3pt} c@{\hskip3pt} @{\hskip3pt}c@{\hskip3pt} @{\hskip3pt} c@{\hskip3pt}  @{\hskip3pt}c@{\hskip3pt}| @{\hskip4pt}c@{\hskip4pt}}
			\toprule
			{\bf Model}             & {Conciseness} & {Fluency} & {No-hallucinations} & {Informativeness} & {\bf Overall} \\ 
			\midrule
			\modelp{UL2}{20B}       & \bf 0.53      & \bf 0.52  & 0.54                & 0.49              & 0.50          \\ 
			\modelp{BART}{LARGE}    & 0.50          & 0.50      & 0.52                & 0.49              & 0.49          \\ 
			\modelp{PEGASUS}{LARGE} & 0.52          & 0.49      & 0.49                & 0.49              & 0.49          \\ 
			\modelp{T5}{11B}        & 0.49          & 0.50      & 0.49                & 0.48              & 0.47          \\ 
			\midrule
			\zcodepplarge{}         & 0.50          & 0.51      & \textbf{0.55}       & \textbf{0.49}     & \textbf{0.51} \\ 
			\bottomrule
		\end{tabular}}
	%\end{adjustbox}
	\caption{
		Human evaluation results on the XSum leaderboard. 
	}
	\label{tab:xsum}
	\vspace{-2mm}
\end{table*}
% \newcolumntype{g}{>{\columncolor{Gray}}c}
% \definecolor{Gray}{gray}{0.95}
% \begin{table*}[htb!]
%     \centering
%     \begin{adjustbox}{max width=0.99\textwidth}
%     \renewcommand{\arraystretch}{1.1}
%     % \begin{tabular}{@{\hskip3pt}l@{\hskip2pt}|@{\hskip2pt} c@{\hskip2pt}|@{\hskip2pt} c@{\hskip2pt}|@{\hskip2pt} c@{\hskip2pt}||@{\hskip2pt} c@{\hskip2pt}|@{\hskip2pt} c@{\hskip2pt}| @{\hskip2pt}c@{\hskip2pt}}
%     \begin{tabular}{lcccccg}
%         \toprule
%         \multirow{2}{*}{\bf Task} & Eval  & Metric  &\modelp{PaLM}{540B}& \modelp{mT5}{xlarge} & \modelp{mT5}{large}& \zcodepplarge{} \\
%         & & & 540B&3B & 705M  & 710M \\ 
%         \midrule
%         \#Training Tokens& & & 500B& 1T& 1T& 500B \\
%        \midrule
%         \multicolumn{7}{c}{Cross-lingual summarization} \\ 
%         \midrule
%         WikiLingua(ru $\rightarrow$ en) &test&R2& 18.6& 14.6 & 11.2 & \bf15.9 \\
%         WikiLingua(vi $\rightarrow$ en) &&& 19.1 & 14.9 & 10.9 & \bf16.7 \\
%         WikiLingua(es $\rightarrow$ en) &&& 20.9 & 17.2 & 12.6 & \bf17.7 \\
%         WikiLingua(tr $\rightarrow$ en) &&& 23.1 & 18.3& 14.5 & \bf22.9 \\ 
%         \midrule
%         \multicolumn{7}{c}{Multi-lingual summarization} \\ 
%         \midrule
%         MLSum(de) &test&R2& 33.1 & 36.2 & 35.4 & \bf36.8 \\
%         MLSum(es) &&& 12.0 & 13.8 & 12.3 & \bf14.8 \\ 
%         \bottomrule
%         \end{tabular}
%         \end{adjustbox}
%     \caption{
%     Comparison results on multi-lingual summarization tasks. 
%     }
%     \label{tab:xlm_sum}
%     \vspace{-2mm}
% \end{table*}

\newcolumntype{g}{>{\columncolor{Gray}}c}
\definecolor{Gray}{gray}{0.97}
\begin{table}[htb!]
    \centering
    %\begin{adjustbox}{max width=0.99\textwidth}
        \renewcommand{\arraystretch}{1.05}
        % \begin{tabular}{@{\hskip3pt}l@{\hskip2pt}|@{\hskip2pt} c@{\hskip2pt}|@{\hskip2pt} c@{\hskip2pt}|@{\hskip2pt} c@{\hskip2pt}||@{\hskip2pt} c@{\hskip2pt}|@{\hskip2pt} c@{\hskip2pt}| @{\hskip2pt}c@{\hskip2pt}}
        \scalebox{0.6}{
        \begin{tabular}{l|c|ccg}
            \toprule
            \multirow{2}{*}{\bf Dataset}        & \modelp{PaLM}{540B} & \modelp{mT5}{XLARGE} & \modelp{mT5}{LARGE} & \zcodepplarge{} \\
                                                    & 540B                & 3B                   & 705M                & 710M            \\ 
            \midrule
            \#Training Tokens                       & 500B                & 1T                   & 1T                  & 500B            \\
            \midrule
            \multicolumn{5}{l}{\emph{Cross-lingual summarization}} \\ 
            \midrule
            WikiLingua (ru $\rightarrow$ en)      & 18.6                & 14.6                 & 11.2                & \bf15.9         \\
            WikiLingua (vi $\rightarrow$ en)         & 19.1                & 14.9                 & 10.9                & \bf16.7         \\
            WikiLingua (es $\rightarrow$ en)         & 20.9                & 17.2                 & 12.6                & \bf17.7         \\
            WikiLingua (tr $\rightarrow$ en)         & 23.1                & 18.3                 & 14.5                & \bf22.9         \\ 

            Average         & 20.4                & 16.3                 & 12.3                & \bf18.3         \\ 
            
            \midrule
            \multicolumn{5}{l}{\emph{Multilingual summarization}} \\ 
            \midrule
            MLSum (de)                           & 33.1                & 36.2                 & 35.4                & \bf36.8         \\
            MLSum (es)                              & 12.0                & 13.8                 & 12.3                & \bf14.8         \\ 
            Average         & 22.6                & 25.0                 & 23.9                & \bf25.8         \\ 
            \bottomrule
        \end{tabular}}
    %\end{adjustbox}
    \caption{
        Evaluation results on multi-lingual summarization tasks. Best numbers excluding \modelp{PaLM}{540B} are in \textbf{Bold}.
    }
    \label{tab:xlm_sum}
    %\vspace{-2mm}
\end{table}
We compare \ModelName{} to \modelp{PEGASUS}{} and \modelp{LongT5}{}, which is optimized for long document summarization. Results in Table \ref{tab:en_long} show that \zcodepplarge{} exceeds all the strong competitors on all long document summarization datasets and lifts SOTA by 0.35 point on average. For FiE, which is used to generate summaries for arXiv and PubMed, we choose the chunk size $l=256$, and choose the last layer of encoder as fusion layer based on the experiment results.
Specifically, \zcodepplarge{} outperforms \modelp{LongT5}{3B} with less than 1/3 of parameters. These results demonstrate both the effectiveness and flexibility of \ModelName{} by using Disentangled-Attention to encode word dependencies.

\subsubsection{Human Evaluation}

As human evaluation is the most reliable measurement of the quality of natural language generation models, we submit the test results of XSum to the leaderboard~\citep{khashabi2021genie} which requires human raters to compare the generated summaries side by side with human written references. Please check the paper of the leaderboad \citep{khashabi2021genie} to get more details of human evaluation process including instructions, dataset preparing, payments and demographics of the raters. We list the human evaluation results in Table\ref{tab:xsum}. \ModelName{} outperforms all the other models, e.g., \modelp{BART}{LARGE}, \modelp{PEGASUS}{LARGE}, \modelp{T5}{11B}, \modelp{UL2}{20B}\citep{tay2022unifying}, on the leaderboard in terms of human-overall score.  As the human evaluation score is an average of side-by-side preference comparison scores, a score of 0.51 indicates that the annotators prefer the output of Z-Code++ to the human written references. Further more, while hallucination is one of the most critical problems for abstractive summarization, Z-Code++ does not suffer much, i.e., 0.55, among the leaderbard. The human evaluation results validate that \ModelName{} produces higher quality summaries than other models.

% It's worth to mention that even though the ROUGE-2 is close to PEGASUS on XSum task, the human evaluation results on XSum leaderboard listed in Table\ref{tab:xsum} indicates a larger gain of \zcodepplarge{} over PEGASUS. We think this is due to the  pre-training objective of PEGASUS, GSG, is specifically biased to ROUGE metric as the gap sentences are selected based on ROUGE-like importance score. 

\subsubsection{Results on Multilingual Summarization}

Following GEM-benchmark~\citep{gehrmann2021gem}, we evaluate the performance of \zcodepplarge{} \footnote{Note that \zcodepplarge{} for multilingual summarization is differently trained. Refer to \ref{sec:setup} for more details.} on multilingual summarization with WikiLingua and MLSum. We compare \zcodepplarge{} with \modelp{mT5}{LARGE} and \modelp{mT5}{XLARGE}. The results of \modelp{PaLM}{540B}, a state of the art PLM, are also listed in Table \ref{tab:xlm_sum}. Compared with \modelp{mT5}{XLARGE}, \zcodepplarge{} achieves substantially better performance across all the tasks with only 1/3 parameters and half training data. In addition, we observe a significant performance gap between \zcodepplarge{} and \modelp{PaLM}{540B} on WikiLingua, which is not surprising due to the sharp difference in model size and capacity. However, \zcodepplarge{} surpasses \modelp{PaLM}{540B} on MLSum by a large margin, i.e., 3.7\% on MLSum(de), 2.8\% on MLSum(es), albeit \zcodepplarge{} has less than 1/500 parameters. We believe that by scaling up \ModelName{} to a moderate size (e.g., 10B), the performance gap on WikiLingua would be mitigated. We leave it to future work.

\subsubsection{Results on Low-Resource Summarization }
We explore how well knowledge learned in different pre-training stages can generalize to low-resource summarization scenarios, \ie{} zero/few-shot evaluation. For the grounded pre-training phase, we choose to include MediaSum, MultiNews, NewsRoom, and WikiHow datasets. Corresponding instructions are listed in Table \ref{tab:grounded_pretraining}. We reckon that incorporating diverse datasets and instructions is beneficial, which we leave it to future work. For the fine-tuning stage, following the setting in~\citet{zhang2020pegasus}, we randomly select the number of training data to 0, 10, 100, and 1000, and sample examples from XSUM, CNNDM, and SAMSum, and then fine-tune \ModelName{} until no significant improvement on the validation set is observed. Note that 0 denotes zero-shot evaluation. Table \ref{tab:fewshot_evaluation} presents the results. By fine-tuning first-phase pre-trained model, \zcodepplarge{} outperforms \modelp{T5}{LARGE} by more than 3 points on average. \modelp{PEGASUS}{LARGE} exceeds \zcodepplarge{} when the number of training examples is less than 100, which is foreseeable as \modelp{PEGASUS}{LARGE} is pre-trained with a pseudo summarization objective. However, \zcodepplarge{} performs significantly better than them when it is trained with more than 100 examples, showing strong generalization in the few-shot setting. More importantly, with grounded pre-training, \zcodepplarge{} beats all the competing models by a large margin in both zero and few-shot settings, outperforming \modelp{PEGASUS}{LARGE} by 5.7/1.5/3.3 points on average. This suggests that instructions-grounded pre-training enables effective knowledge transfer to downstream low-resource tasks.

\begin{table}[htb!]
	\centering
	%\begin{adjustbox}{max width=0.99\textwidth}
 \renewcommand{\arraystretch}{1.05}
		% \begin{tabular}{@{\hskip3pt}l@{\hskip3pt}|@{\hskip10pt} c@{\hskip20pt} @{\hskip3pt}c@{\hskip25pt} @{\hskip3pt}c@{\hskip25pt} @{\hskip3pt} c@{\hskip4pt} | @{\hskip6pt} c@{\hskip10pt}}
  \scalebox{0.8}{
  \begin{tabular}{l|cccc|c}
			\toprule
			{\bf Model}                & {0}         & {10}        & {100}    & {1000} & {\textbf{Average}}  \\ 
			\midrule
			&
			\multicolumn{5}{c}{XSUM} \\ \midrule
			\modelp{T5}{LARGE}         & 2.3         & 2.5         & 5.5      & 9.4   & 4.9  \\
			\modelp{PEGASUS}{LARGE}    & 3.0         & 3.5         & 16.4     & 18.2  & 10.3  \\
			\zcodepp$_{\mathtt{LARGE}}^{\dag}$          & 0.1         & 2.1         & 12.3     & 17.3  & 8.0  \\ 
			\zcodepplarge{} & \bf13.7     & \bf14.0     & \bf17.5  & \bf18.9 & \bf16.0
			% {\ModelName}\textsubscript{large}  &  \bf13.7  &  \bf14.0  &  \bf14.6 & \bf18.9 
			\\ \midrule
			&
			\multicolumn{5}{c}{CNNDM} \\ \midrule
			\modelp{T5}{LARGE}         & 4.9         & 5.1         & 7.7      & 11.2  & 2.7  \\
			\modelp{PEGASUS}{LARGE}    & 13.3        & 15.8        & 18.2     & 19.4  & 16.7  \\
			%  GPT3.5(7-shot) & -& 9.19& -&- \\
			%  InstGPT3.5(7-shot) & -& 15.90&- &- \\
			\zcodepp$_{\mathtt{LARGE}}^{\dag}$            & 0.1         & 1.5         & 15.0     & 18.3  & 8.7  \\
			\zcodepplarge{} & \bf17.3\rlap{$^*$} & \bf17.3\rlap{$^*$} & \bf18.4  & \bf19.6 & \bf18.2 \\
			% {\ModelName}\textsubscript{large} &  \bf17.3$^*$  &  \bf17.3$^*$  &  \bf17.3$^*$ & \bf18.7 \\
			\midrule
			&
			\multicolumn{5}{c}{SAMSum} \\ \midrule
			\modelp{T5}{LARGE}         & 1.3         & 4.0         & 10.4     & 17.8   & 8.4 \\
			\modelp{PEGASUS}{LARGE}    & 6.4         & 11.7        & 19.8     & 24.4   & 15.6 \\
			\zcodepp$_{\mathtt{LARGE}}^{\dag}$            & 0.1         & 2.6         & 20.2     & 26.3   & 12.3 \\
			\zcodepplarge{} & \bf7.9      & \bf17.4     & \bf 22.3 & \bf28.1 & \bf18.9 \\
			% {\ModelName}\textsubscript{large} &  \bf7.9  &  \bf17.4  &  \bf 22.3 & \bf28.1  \\
			\bottomrule
		\end{tabular}}
	%\end{adjustbox} 
	\caption{ROUGE-2 score in different summarization datasets. Results are shown on their full test sets using 10, 100, and 1000 training examples. 0 denotes zero-shot results. Results marked with $^*$ mean that unfine-tuned checkpoints perform the best, i.e., zero-shot performance is better than the fine-tuned one. \zcodepp$_{\mathtt{LARGE}}^{\dag}$ refers to fine-tuning from phase 1 pre-trained model. \zcodepplarge{} fine-tuned from two-phase pre-trained model. }
	\label{tab:fewshot_evaluation}
	\vspace{-1mm}
\end{table}

% which indicates the necessity of \emph{grounded pre-training} on generalization to low-resource scenarios. A few examples are provided in appendix to demonstrate 
%In real-world practice, it hard to collect training data to adapt the pre-trained model to a new domain or task. It's important to measure the Zeroshot and Fewshot performance of foundation model. To measure few-shot performance, we randomly sample 10,100 samples from XSum, CNNDM and SAMSum to fine-tune Z-Code++ and measure it's performance on the test datasets of corresponding tasks. Following T0, we also jointly train Z-Code++ model with multiple Summarizationt tasks, including WikiHow, Newsroom, MediaSum and MultiNews. We prepend a diverse prompts with training documents of different tasks. 

% \input{floats/results_fewshot_summarization_full}

\section{Conclusions}
\label{sec:conclusion}

We present Z-Code++, an efficient and effective pre-trained language model optimized for abstractive text summarization. 
The model extends the encoder-decoder model using three techniques. The first is a two-phase pre-training process, where the model is first pre-trained using text corpora for language understanding, and then is continually pre-trained on summarization corpora for grounded text generation. The second technique is the use of the disentangled attention mechanism, where each word is represented using two vectors that encode its content and position, respectively. The third is the fusion-in-encoder method for encoding long sequence inputs.
We present a comprehensive empirical study to validate the effectiveness of Z-Code++.
The model creates new state of the art on 9 out of 13 text summarization tasks across 5 languages. 
In addition, we show that our model is parameter-efficient in that it outperforms the 600x larger PaLM\textsubscript{540B} on XSum, and the finetuned 200x larger GPT3\textsubscript{175B} on SAMSum. 
Z-Code++ also generalizes well to low-resource downstream tasks. For example, in zero-shot and few-shot settings, our model outperforms more substantially the competing models.

However, evaluation \citep{liang2022holistic} and hallucinations are still two long-standing problems of summarizations that we do not touch with in this work, in the future we will 1) explore evaluation metrics that correlate well with human experience, 2) learn to summarize to better align with human preferences \citep{stiennon2020learning, ouyang2022training}, and 3) ground summarization models on world knowledge to largely reduce hallucinations \citep{lecun2022path, hafner2023mastering}.
\section*{Limitations}
In this paper, we introduce \zcodepp{}, a robust pre-trained model tailored for summarization tasks. However, it should be noted that there are certain limitations to our model. Firstly, the model is not versatile enough as it is specifically designed for summarization. It is unclear whether it performs well on other natural language tasks. Secondly, while FiE can handle document summarization, there are still significant potential for improving cost efficiency. Lastly, the evaluation of multilingual summarization is not thorough enough due to the limitations of available datasets. We intend to address these limitations in our future work.

%the encoder of \zcodepp{} does not possess a deep understanding of the document as it is not explicitly designed to comprehend it. We plan to address these limitations in our future work.
\section*{Ethics Statement}
The same as all existing generative language models, the generated text of \zcodepp{} raises various ethical considerations. One crucial consideration is the issue of potential hallucinations in the summaries generated by the model. The summaries produced by a generative model may not necessarily be faithful to the original article or entirely factual which may mislead the users to make incorrect decisions based on the summary without additional knowledge. In addition, another important consideration is the potential for bias in generated summaries, such as bias based on gender, race, and other factors.

\bibliography{ref,ref_xiaodong}

\begin{thebibliography}{61}
\expandafter\ifx\csname natexlab\endcsname\relax\def\natexlab#1{#1}\fi

\bibitem[{Aghajanyan et~al.(2021)Aghajanyan, Gupta, Shrivastava, Chen,
  Zettlemoyer, and Gupta}]{aghajanyan2021muppet}
Armen Aghajanyan, Anchit Gupta, Akshat Shrivastava, Xilun Chen, Luke
  Zettlemoyer, and Sonal Gupta. 2021.
\newblock Muppet: Massive multi-task representations with pre-finetuning.
\newblock \emph{arXiv preprint arXiv:2101.11038}.

\bibitem[{Allahyari et~al.(2017)Allahyari, Pouriyeh, Assefi, Safaei, Trippe,
  Gutierrez, and Kochut}]{allahyari2017text}
Mehdi Allahyari, Seyedamin Pouriyeh, Mehdi Assefi, Saeid Safaei, Elizabeth~D
  Trippe, Juan~B Gutierrez, and Krys Kochut. 2017.
\newblock Text summarization techniques: a brief survey.
\newblock \emph{arXiv preprint arXiv:1707.02268}.

\bibitem[{Bajaj et~al.(2022)Bajaj, Xiong, Ke, Liu, He, Tiwary, Liu, Bennett,
  Song, and Gao}]{bajaj2022metro}
Payal Bajaj, Chenyan Xiong, Guolin Ke, Xiaodong Liu, Di~He, Saurabh Tiwary,
  Tie-Yan Liu, Paul Bennett, Xia Song, and Jianfeng Gao. 2022.
\newblock Metro: Efficient denoising pretraining of large scale autoencoding
  language models with model generated signals.
\newblock \emph{arXiv preprint arXiv:2204.06644}.

\bibitem[{Brown et~al.(2020)Brown, Mann, Ryder, Subbiah, Kaplan, Dhariwal,
  Neelakantan, Shyam, Sastry, Askell et~al.}]{gpt3}
Tom~B Brown, Benjamin Mann, Nick Ryder, Melanie Subbiah, Jared Kaplan, Prafulla
  Dhariwal, Arvind Neelakantan, Pranav Shyam, Girish Sastry, Amanda Askell,
  et~al. 2020.
\newblock Language models are few-shot learners.
\newblock \emph{arXiv preprint arXiv:2005.14165}.

\bibitem[{Celikyilmaz et~al.(2020)Celikyilmaz, Clark, and
  Gao}]{celikyilmaz2020evaluation}
Asli Celikyilmaz, Elizabeth Clark, and Jianfeng Gao. 2020.
\newblock Evaluation of text generation: A survey.
\newblock \emph{arXiv preprint arXiv:2006.14799}.

\bibitem[{Chopra et~al.(2016)Chopra, Auli, and Rush}]{chopra2016abstractive}
Sumit Chopra, Michael Auli, and Alexander~M Rush. 2016.
\newblock Abstractive sentence summarization with attentive recurrent neural
  networks.
\newblock In \emph{Proceedings of the 2016 Conference of the North American
  Chapter of the Association for Computational Linguistics: Human Language
  Technologies}, pages 93--98.

\bibitem[{Chowdhery et~al.(2022)Chowdhery, Narang, Devlin, Bosma, Mishra,
  Roberts, Barham, Chung, Sutton, Gehrmann et~al.}]{chowdhery2022palm}
Aakanksha Chowdhery, Sharan Narang, Jacob Devlin, Maarten Bosma, Gaurav Mishra,
  Adam Roberts, Paul Barham, Hyung~Won Chung, Charles Sutton, Sebastian
  Gehrmann, et~al. 2022.
\newblock Palm: Scaling language modeling with pathways.
\newblock \emph{arXiv preprint arXiv:2204.02311}.

\bibitem[{Clark et~al.(2020)Clark, Luong, Le, and Manning}]{clark2020electra}
Kevin Clark, Minh-Thang Luong, Quoc~V. Le, and Christopher~D. Manning. 2020.
\newblock {ELECTRA}: Pre-training text encoders as discriminators rather than
  generators.
\newblock In \emph{ICLR}.

\bibitem[{Cohan et~al.(2018)Cohan, Dernoncourt, Kim, Bui, Kim, Chang, and
  Goharian}]{cohan2018discourse}
Arman Cohan, Franck Dernoncourt, Doo~Soon Kim, Trung Bui, Seokhwan Kim, Walter
  Chang, and Nazli Goharian. 2018.
\newblock A discourse-aware attention model for abstractive summarization of
  long documents.
\newblock \emph{arXiv preprint arXiv:1804.05685}.

\bibitem[{El-Kishky et~al.(2019)El-Kishky, Chaudhary, Guzm{\'a}n, and
  Koehn}]{el2019ccaligned}
Ahmed El-Kishky, Vishrav Chaudhary, Francisco Guzm{\'a}n, and Philipp Koehn.
  2019.
\newblock Ccaligned: A massive collection of cross-lingual web-document pairs.
\newblock \emph{arXiv preprint arXiv:1911.06154}.

\bibitem[{Fabbri et~al.(2019)Fabbri, Li, She, Li, and Radev}]{fabbri2019multi}
Alexander~R Fabbri, Irene Li, Tianwei She, Suyi Li, and Dragomir~R Radev. 2019.
\newblock Multi-news: A large-scale multi-document summarization dataset and
  abstractive hierarchical model.
\newblock \emph{arXiv preprint arXiv:1906.01749}.

\bibitem[{Gao et~al.(2022)Gao, Xiong, Bennett, and Craswell}]{gao2022neural}
Jianfeng Gao, Chenyan Xiong, Paul Bennett, and Nick Craswell. 2022.
\newblock Neural approaches to conversational information retrieval.
\newblock \emph{arXiv preprint arXiv:2201.05176}.

\bibitem[{Gehrmann et~al.(2021)Gehrmann, Adewumi, Aggarwal, Ammanamanchi,
  Anuoluwapo, Bosselut, Chandu, Clinciu, Das, Dhole et~al.}]{gehrmann2021gem}
Sebastian Gehrmann, Tosin Adewumi, Karmanya Aggarwal, Pawan~Sasanka
  Ammanamanchi, Aremu Anuoluwapo, Antoine Bosselut, Khyathi~Raghavi Chandu,
  Miruna Clinciu, Dipanjan Das, Kaustubh~D Dhole, et~al. 2021.
\newblock The gem benchmark: Natural language generation, its evaluation and
  metrics.
\newblock \emph{arXiv preprint arXiv:2102.01672}.

\bibitem[{Gehrmann et~al.(2022)Gehrmann, Clark, and
  Sellam}]{gehrmann2022repairing}
Sebastian Gehrmann, Elizabeth Clark, and Thibault Sellam. 2022.
\newblock Repairing the cracked foundation: A survey of obstacles in evaluation
  practices for generated text.
\newblock \emph{arXiv preprint arXiv:2202.06935}.

\bibitem[{Gliwa et~al.(2019)Gliwa, Mochol, Biesek, and Wawer}]{gliwa2019samsum}
Bogdan Gliwa, Iwona Mochol, Maciej Biesek, and Aleksander Wawer. 2019.
\newblock Samsum corpus: A human-annotated dialogue dataset for abstractive
  summarization.
\newblock \emph{arXiv preprint arXiv:1911.12237}.

\bibitem[{Grusky et~al.(2018)Grusky, Naaman, and Artzi}]{grusky2018newsroom}
Max Grusky, Mor Naaman, and Yoav Artzi. 2018.
\newblock Newsroom: A dataset of 1.3 million summaries with diverse extractive
  strategies.
\newblock \emph{arXiv preprint arXiv:1804.11283}.

\bibitem[{Guo et~al.(2021)Guo, Ainslie, Uthus, Ontanon, Ni, Sung, and
  Yang}]{guo2021longt5}
Mandy Guo, Joshua Ainslie, David Uthus, Santiago Ontanon, Jianmo Ni, Yun-Hsuan
  Sung, and Yinfei Yang. 2021.
\newblock Longt5: Efficient text-to-text transformer for long sequences.
\newblock \emph{arXiv preprint arXiv:2112.07916}.

\bibitem[{Hafner et~al.(2023)Hafner, Pasukonis, Ba, and
  Lillicrap}]{hafner2023mastering}
Danijar Hafner, Jurgis Pasukonis, Jimmy Ba, and Timothy Lillicrap. 2023.
\newblock Mastering diverse domains through world models.
\newblock \emph{arXiv preprint arXiv:2301.04104}.

\bibitem[{Hao et~al.(2021)Hao, Dong, Bao, Xu, and Wei}]{hao2021learning}
Yaru Hao, Li~Dong, Hangbo Bao, Ke~Xu, and Furu Wei. 2021.
\newblock Learning to sample replacements for electra pre-training.
\newblock \emph{arXiv preprint arXiv:2106.13715}.

\bibitem[{He et~al.(2021)He, Gao, and Chen}]{he2021debertav3}
Pengcheng He, Jianfeng Gao, and Weizhu Chen. 2021.
\newblock Debertav3: Improving deberta using electra-style pre-training with
  gradient-disentangled embedding sharing.
\newblock \emph{arXiv preprint arXiv:2111.09543}.

\bibitem[{He et~al.(2020)He, Liu, Gao, and Chen}]{he2020deberta}
Pengcheng He, Xiaodong Liu, Jianfeng Gao, and Weizhu Chen. 2020.
\newblock Deberta: Decoding-enhanced bert with disentangled attention.
\newblock In \emph{International Conference on Learning Representations}.

\bibitem[{Hu et~al.(2021)Hu, Shen, Wallis, Allen-Zhu, Li, Wang, Wang, and
  Chen}]{hu2021lora}
Edward~J Hu, Yelong Shen, Phillip Wallis, Zeyuan Allen-Zhu, Yuanzhi Li, Shean
  Wang, Lu~Wang, and Weizhu Chen. 2021.
\newblock Lora: Low-rank adaptation of large language models.
\newblock \emph{arXiv preprint arXiv:2106.09685}.

\bibitem[{Izacard and Grave(2020)}]{izacard2020leveraging}
Gautier Izacard and Edouard Grave. 2020.
\newblock Leveraging passage retrieval with generative models for open domain
  question answering.
\newblock \emph{arXiv preprint arXiv:2007.01282}.

\bibitem[{Ji et~al.(2022)Ji, Lee, Frieske, Yu, Su, Xu, Ishii, Bang, Madotto,
  and Fung}]{ji2022survey}
Ziwei Ji, Nayeon Lee, Rita Frieske, Tiezheng Yu, Dan Su, Yan Xu, Etsuko Ishii,
  Yejin Bang, Andrea Madotto, and Pascale Fung. 2022.
\newblock Survey of hallucination in natural language generation.
\newblock \emph{arXiv preprint arXiv:2202.03629}.

\bibitem[{Joshi et~al.(2020)Joshi, Chen, Liu, Weld, Zettlemoyer, and
  Levy}]{joshi2019spanbert}
Mandar Joshi, Danqi Chen, Yinhan Liu, Daniel~S Weld, Luke Zettlemoyer, and Omer
  Levy. 2020.
\newblock Spanbert: Improving pre-training by representing and predicting
  spans.
\newblock \emph{Transactions of the Association for Computational Linguistics},
  8:64--77.

\bibitem[{Khashabi et~al.(2021)Khashabi, Stanovsky, Bragg, Lourie, Kasai, Choi,
  Smith, and Weld}]{khashabi2021genie}
Daniel Khashabi, Gabriel Stanovsky, Jonathan Bragg, Nicholas Lourie, Jungo
  Kasai, Yejin Choi, Noah~A Smith, and Daniel~S Weld. 2021.
\newblock Genie: A leaderboard for human-in-the-loop evaluation of text
  generation.
\newblock \emph{arXiv preprint arXiv:2101.06561}.

\bibitem[{Kim et~al.(2018)Kim, Kim, and Kim}]{kim2018abstractive}
Byeongchang Kim, Hyunwoo Kim, and Gunhee Kim. 2018.
\newblock Abstractive summarization of reddit posts with multi-level memory
  networks.
\newblock \emph{arXiv preprint arXiv:1811.00783}.

\bibitem[{Kry{\'s}ci{\'n}ski et~al.(2019)Kry{\'s}ci{\'n}ski, Keskar, McCann,
  Xiong, and Socher}]{kryscinski2019neural}
Wojciech Kry{\'s}ci{\'n}ski, Nitish~Shirish Keskar, Bryan McCann, Caiming
  Xiong, and Richard Socher. 2019.
\newblock Neural text summarization: A critical evaluation.
\newblock In \emph{Proceedings of the 2019 Conference on Empirical Methods in
  Natural Language Processing and the 9th International Joint Conference on
  Natural Language Processing (EMNLP-IJCNLP)}, pages 540--551.

\bibitem[{Ladhak et~al.(2020)Ladhak, Durmus, Cardie, and
  McKeown}]{ladhak2020wikilingua}
Faisal Ladhak, Esin Durmus, Claire Cardie, and Kathleen McKeown. 2020.
\newblock Wikilingua: A new benchmark dataset for cross-lingual abstractive
  summarization.
\newblock \emph{arXiv preprint arXiv:2010.03093}.

\bibitem[{Lample and Conneau(2019)}]{lample2019xlm}
Guillaume Lample and Alexis Conneau. 2019.
\newblock Cross-lingual language model pretraining.
\newblock \emph{NeurIPS}.

\bibitem[{LeCun(2022)}]{lecun2022path}
Yann LeCun. 2022.
\newblock A path towards autonomous machine intelligence version 0.9. 2,
  2022-06-27.
\newblock \emph{Open Review}, 62.

\bibitem[{Liang et~al.(2022)Liang, Bommasani, Lee, Tsipras, Soylu, Yasunaga,
  Zhang, Narayanan, Wu, Kumar et~al.}]{liang2022holistic}
Percy Liang, Rishi Bommasani, Tony Lee, Dimitris Tsipras, Dilara Soylu,
  Michihiro Yasunaga, Yian Zhang, Deepak Narayanan, Yuhuai Wu, Ananya Kumar,
  et~al. 2022.
\newblock Holistic evaluation of language models.
\newblock \emph{arXiv preprint arXiv:2211.09110}.

\bibitem[{Liu and Lapata(2019{\natexlab{a}})}]{liu2019hierarchical}
Yang Liu and Mirella Lapata. 2019{\natexlab{a}}.
\newblock Hierarchical transformers for multi-document summarization.
\newblock In \emph{Proceedings of the 57th Annual Meeting of the Association
  for Computational Linguistics}, pages 5070--5081.

\bibitem[{Liu and Lapata(2019{\natexlab{b}})}]{liu2019text}
Yang Liu and Mirella Lapata. 2019{\natexlab{b}}.
\newblock Text summarization with pretrained encoders.
\newblock In \emph{Proceedings of the 2019 Conference on Empirical Methods in
  Natural Language Processing and the 9th International Joint Conference on
  Natural Language Processing (EMNLP-IJCNLP)}, pages 3730--3740.

\bibitem[{Nallapati et~al.(2016)Nallapati, Zhou, dos Santos, Gulcehre, and
  Xiang}]{nallapati2016abstractive}
Ramesh Nallapati, Bowen Zhou, Cicero dos Santos, Caglar Gulcehre, and Bing
  Xiang. 2016.
\newblock Abstractive text summarization using sequence-to-sequence rnns and
  beyond.
\newblock In \emph{Proceedings of The 20th SIGNLL Conference on Computational
  Natural Language Learning}, pages 280--290.

\bibitem[{Narayan et~al.(2018)Narayan, Cohen, and Lapata}]{narayan2018don}
Shashi Narayan, Shay~B Cohen, and Mirella Lapata. 2018.
\newblock Don't give me the details, just the summary! topic-aware
  convolutional neural networks for extreme summarization.
\newblock \emph{arXiv preprint arXiv:1808.08745}.

\bibitem[{Ouyang et~al.(2022)Ouyang, Wu, Jiang, Almeida, Wainwright, Mishkin,
  Zhang, Agarwal, Slama, Ray et~al.}]{ouyang2022training}
Long Ouyang, Jeff Wu, Xu~Jiang, Diogo Almeida, Carroll~L Wainwright, Pamela
  Mishkin, Chong Zhang, Sandhini Agarwal, Katarina Slama, Alex Ray, et~al.
  2022.
\newblock Training language models to follow instructions with human feedback.
\newblock \emph{arXiv preprint arXiv:2203.02155}.

\bibitem[{Pang et~al.(2022)Pang, Nijkamp, Kry{\'s}ci{\'n}ski, Savarese, Zhou,
  and Xiong}]{pang2022long}
Bo~Pang, Erik Nijkamp, Wojciech Kry{\'s}ci{\'n}ski, Silvio Savarese, Yingbo
  Zhou, and Caiming Xiong. 2022.
\newblock Long document summarization with top-down and bottom-up inference.
\newblock \emph{arXiv preprint arXiv:2203.07586}.

\bibitem[{Pasunuru et~al.(2021)Pasunuru, Celikyilmaz, Galley, Xiong, Zhang,
  Bansal, and Gao}]{pasunuru2021data}
Ramakanth Pasunuru, Asli Celikyilmaz, Michel Galley, Chenyan Xiong, Yizhe
  Zhang, Mohit Bansal, and Jianfeng Gao. 2021.
\newblock Data augmentation for abstractive query-focused multi-document
  summarization.
\newblock In \emph{Proceedings of the AAAI Conference on Artificial
  Intelligence}, volume~35, pages 13666--13674.

\bibitem[{Peng et~al.(2022)Peng, Galley, He, Brockett, Liden, Nouri, Yu, Dolan,
  and Gao}]{GODEL}
Baolin Peng, Michel Galley, Pengcheng He, Chris Brockett, Lars Liden, Elnaz
  Nouri, Zhou Yu, Bill Dolan, and Jianfeng Gao. 2022.
\newblock Large-scale pre-training for goal-directed dialogue.
\newblock Technical report, Microsoft Technical Report.

\bibitem[{Radford et~al.(2019)Radford, Wu, Child, Luan, Amodei, and
  Sutskever}]{gpt2}
Alec Radford, Jeffrey Wu, Rewon Child, David Luan, Dario Amodei, and Ilya
  Sutskever. 2019.
\newblock Language models are unsupervised multitask learners.
\newblock \emph{OpenAI Blog}, 1(8).

\bibitem[{Raffel et~al.(2020)Raffel, Shazeer, Roberts, Lee, Narang, Matena,
  Zhou, Li, and Liu}]{raffel2019t5}
Colin Raffel, Noam Shazeer, Adam Roberts, Katherine Lee, Sharan Narang, Michael
  Matena, Yanqi Zhou, Wei Li, and Peter~J. Liu. 2020.
\newblock \href {http://jmlr.org/papers/v21/20-074.html} {Exploring the limits
  of transfer learning with a unified text-to-text transformer}.
\newblock \emph{Journal of Machine Learning Research}, 21(140):1--67.

\bibitem[{Rajpurkar et~al.(2016)Rajpurkar, Zhang, Lopyrev, and Liang}]{squad1}
Pranav Rajpurkar, Jian Zhang, Konstantin Lopyrev, and Percy Liang. 2016.
\newblock {SQ}u{AD}: 100,000+ questions for machine comprehension of text.
\newblock In \emph{Proceedings of the 2016 Conference on Empirical Methods in
  Natural Language Processing}.

\bibitem[{Rothe et~al.(2021)Rothe, Maynez, and Narayan}]{rothe2021thorough}
Sascha Rothe, Joshua Maynez, and Shashi Narayan. 2021.
\newblock A thorough evaluation of task-specific pretraining for summarization.
\newblock In \emph{Proceedings of the 2021 Conference on Empirical Methods in
  Natural Language Processing}, pages 140--145.

\bibitem[{Rush et~al.(2015)Rush, Chopra, and Weston}]{rush2015neural}
Alexander~M Rush, Sumit Chopra, and Jason Weston. 2015.
\newblock A neural attention model for abstractive sentence summarization.
\newblock In \emph{Proceedings of the 2015 Conference on Empirical Methods in
  Natural Language Processing}, pages 379--389.

\bibitem[{Sanh et~al.(2022)Sanh, Webson, Raffel, Bach, Sutawika, Alyafeai,
  Chaffin, Stiegler, Le~Scao, Raja et~al.}]{sanh2022multitask}
Victor Sanh, Albert Webson, Colin Raffel, Stephen Bach, Lintang Sutawika, Zaid
  Alyafeai, Antoine Chaffin, Arnaud Stiegler, Teven Le~Scao, Arun Raja, et~al.
  2022.
\newblock Multitask prompted training enables zero-shot task generalization.
\newblock In \emph{The Tenth International Conference on Learning
  Representations}.

\bibitem[{Schwenk et~al.(2019)Schwenk, Wenzek, Edunov, Grave, and
  Joulin}]{schwenk2019ccmatrix}
Holger Schwenk, Guillaume Wenzek, Sergey Edunov, Edouard Grave, and Armand
  Joulin. 2019.
\newblock Ccmatrix: Mining billions of high-quality parallel sentences on the
  web.
\newblock \emph{arXiv preprint arXiv:1911.04944}.

\bibitem[{Scialom et~al.(2020)Scialom, Dray, Lamprier, Piwowarski, and
  Staiano}]{scialom2020mlsum}
Thomas Scialom, Paul-Alexis Dray, Sylvain Lamprier, Benjamin Piwowarski, and
  Jacopo Staiano. 2020.
\newblock Mlsum: The multilingual summarization corpus.
\newblock \emph{arXiv preprint arXiv:2004.14900}.

\bibitem[{See et~al.(2017)See, Liu, and Manning}]{see2017get}
Abigail See, Peter~J Liu, and Christopher~D Manning. 2017.
\newblock Get to the point: Summarization with pointer-generator networks.
\newblock \emph{arXiv preprint arXiv:1704.04368}.

\bibitem[{Stiennon et~al.(2020)Stiennon, Ouyang, Wu, Ziegler, Lowe, Voss,
  Radford, Amodei, and Christiano}]{stiennon2020learning}
Nisan Stiennon, Long Ouyang, Jeffrey Wu, Daniel Ziegler, Ryan Lowe, Chelsea
  Voss, Alec Radford, Dario Amodei, and Paul~F Christiano. 2020.
\newblock Learning to summarize with human feedback.
\newblock \emph{Advances in Neural Information Processing Systems},
  33:3008--3021.

\bibitem[{Tay et~al.(2022)Tay, Dehghani, Tran, Garcia, Bahri, Schuster, Zheng,
  Houlsby, and Metzler}]{tay2022unifying}
Yi~Tay, Mostafa Dehghani, Vinh~Q Tran, Xavier Garcia, Dara Bahri, Tal Schuster,
  Huaixiu~Steven Zheng, Neil Houlsby, and Donald Metzler. 2022.
\newblock Unifying language learning paradigms.
\newblock \emph{arXiv preprint arXiv:2205.05131}.

\bibitem[{Wang et~al.(2019)Wang, Singh, Michael, Hill, Levy, and
  Bowman}]{wang2018glue}
Alex Wang, Amanpreet Singh, Julian Michael, Felix Hill, Omer Levy, and Samuel
  Bowman. 2019.
\newblock Glue: A multi-task benchmark and analysis platform for natural
  language understanding.
\newblock In \emph{7th International Conference on Learning Representations,
  ICLR 2019}.

\bibitem[{Wang et~al.(2020)Wang, Zhai, and Hassan}]{wang2020multi}
Yiren Wang, ChengXiang Zhai, and Hany Hassan. 2020.
\newblock Multi-task learning for multilingual neural machine translation.
\newblock In \emph{Proceedings of the 2020 Conference on Empirical Methods in
  Natural Language Processing (EMNLP)}, pages 1022--1034.

\bibitem[{Wei et~al.(2021)Wei, Bosma, Zhao, Guu, Yu, Lester, Du, Dai, and
  Le}]{wei2021finetuned}
Jason Wei, Maarten Bosma, Vincent~Y Zhao, Kelvin Guu, Adams~Wei Yu, Brian
  Lester, Nan Du, Andrew~M Dai, and Quoc~V Le. 2021.
\newblock Finetuned language models are zero-shot learners.
\newblock \emph{arXiv preprint arXiv:2109.01652}.

\bibitem[{Williams et~al.(2018)Williams, Nangia, and Bowman}]{mnli2018}
Adina Williams, Nikita Nangia, and Samuel Bowman. 2018.
\newblock \href {http://aclweb.org/anthology/N18-1101} {A broad-coverage
  challenge corpus for sentence understanding through inference}.
\newblock In \emph{Proceedings of the 2018 Conference of the North American
  Chapter of the Association for Computational Linguistics: Human Language
  Technologies, Volume 1 (Long Papers)}, pages 1112--1122. Association for
  Computational Linguistics.

\bibitem[{Xiao et~al.(2021)Xiao, Beltagy, Carenini, and Cohan}]{xiao2021primer}
Wen Xiao, Iz~Beltagy, Giuseppe Carenini, and Arman Cohan. 2021.
\newblock Primer: Pyramid-based masked sentence pre-training for multi-document
  summarization.
\newblock \emph{arXiv preprint arXiv:2110.08499}.

\bibitem[{Xue et~al.(2021)Xue, Constant, Roberts, Kale, Al-Rfou, Siddhant,
  Barua, and Raffel}]{xue2021mt5}
Linting Xue, Noah Constant, Adam Roberts, Mihir Kale, Rami Al-Rfou, Aditya
  Siddhant, Aditya Barua, and Colin Raffel. 2021.
\newblock mt5: A massively multilingual pre-trained text-to-text transformer.
\newblock In \emph{Proceedings of the 2021 Conference of the North American
  Chapter of the Association for Computational Linguistics: Human Language
  Technologies}, pages 483--498.

\bibitem[{Zhang et~al.(2020)Zhang, Zhao, Saleh, and Liu}]{zhang2020pegasus}
Jingqing Zhang, Yao Zhao, Mohammad Saleh, and Peter Liu. 2020.
\newblock Pegasus: Pre-training with extracted gap-sentences for abstractive
  summarization.
\newblock In \emph{International Conference on Machine Learning}, pages
  11328--11339. PMLR.

\bibitem[{Zhang et~al.(2021)Zhang, Negrinho, Ghosh, Jagannathan, Hassanzadeh,
  Schaaf, and Gormley}]{zhang2021leveraging}
Longxiang Zhang, Renato Negrinho, Arindam Ghosh, Vasudevan Jagannathan,
  Hamid~Reza Hassanzadeh, Thomas Schaaf, and Matthew~R Gormley. 2021.
\newblock Leveraging pretrained models for automatic summarization of
  doctor-patient conversations.
\newblock In \emph{Findings of the Association for Computational Linguistics:
  EMNLP 2021}, pages 3693--3712.

\bibitem[{Zhu et~al.(2021)Zhu, Liu, Mei, and Zeng}]{zhu2021mediasum}
Chenguang Zhu, Yang Liu, Jie Mei, and Michael Zeng. 2021.
\newblock Mediasum: A large-scale media interview dataset for dialogue
  summarization.
\newblock \emph{arXiv preprint arXiv:2103.06410}.

\bibitem[{Zoph et~al.(2022)Zoph, Bello, Kumar, Du, Huang, Dean, Shazeer, and
  Fedus}]{zoph2022designing}
Barret Zoph, Irwan Bello, Sameer Kumar, Nan Du, Yanping Huang, Jeff Dean, Noam
  Shazeer, and William Fedus. 2022.
\newblock Designing effective sparse expert models.
\newblock \emph{arXiv preprint arXiv:2202.08906}.

\end{thebibliography}
\bibliographystyle{acl_natbib.bst}
\clearpage
\appendix
\section{Appendix}
\label{sec:appendix}

\subsection{Hyper parameters}
\label{grid}

\begin{table}[htb!]
    \centering
    \scalebox{0.85}{
    \begin{tabular}{@{\hskip3pt}l@{\hskip2pt}|@{\hskip2pt} c@{\hskip2pt}}
        \toprule
          Hyper-parameter & Z-Code++\textsubscript{LARGE}\\
        \midrule
        Warmup Steps & \{50,100,500,1000,1500\}\\
        Learning Rates & \{5e-6, 8e-6, 9e-6, 1e-5\}\\
        Batch Size & \{16,32,64\}\\
        Weight Decay & 0.01 \\
        Maximun Training Epochs & \{10,20\}\\
        Learning Rate Decay & Linear \\
        Adam $\epsilon$ & 1e-6 \\
        Adam $\beta_1$ & 0.9 \\
        Adam $\beta_2$ & 0.999 \\
        Gradient Clipping & 1.0  \\ \hline
        Beam search size & \{2,4,5,8\} \\
        Length penalty & \{0.5-1.2\} \\
        Repeated nGram blocking & \{0,3\} \\
        \bottomrule
    \end{tabular}
    }
    \caption{
    Hyper-parameters for fine-tuning Z-Code++ on summarization tasks. 
    }
     \label{tbl:hyper-ft}
\end{table}

\begin{table}[htb!]
  \centering
  \scalebox{0.85}{
    \begin{tabular}{@{\hskip3pt}l@{\hskip2pt}|@{\hskip2pt} c@{\hskip2pt}}
      \toprule
      Hyper-parameter         & Z-Code++\textsubscript{LARGE} \\
      \midrule
      Warmup Steps            & \{1500\}      \\
      Learning Rates          & \{5e-6, 1e-5, 2e-6\}    \\
      Batch Size              & \{64\}                  \\
      Weight Decay            & 0.01                          \\
      Maximun Training Epochs & \{10,20\}                     \\
      Learning Rate Decay     & Linear                        \\
      Adam $\epsilon$         & 1e-6                          \\
      Adam $\beta_1$          & 0.9                           \\
      Adam $\beta_2$          & 0.999                         \\
      Gradient Clipping       & 1.0                           \\ \hline
      Beam search size        & \{5,8\}                   \\
      Length penalty          & \{0.5-1.2\}                   \\
      Repeated nGram blocking & \{0,3\}                       \\
      \bottomrule
    \end{tabular}
  }
  \caption{
    Hyper-parameters for Z-Code++ grounded training. 
  }
  \label{tbl:hyper-grounded}
\end{table}

%\clearpage
%\subsection{Datasets}

\subsection{Rouge scores of summarization tasks}
We list the rouge scores of summarizaiton tasks in table\ref{tab:rg}
\label{rouge}
\begin{table}[htb!]
	\centering
	\scalebox{0.85}{
 \renewcommand{\arraystretch}{1.05}
	\begin{tabular}{l|c|ccc}
 % {@{\hskip3pt}l@{\hskip2pt}|@{\hskip2pt}c@{\hskip2pt}|@{\hskip2pt}c@{\hskip2pt}|@{\hskip2pt}c@{\hskip2pt}|@{\hskip2pt}c@{\hskip2pt}}
		\toprule
		{\bf Task} & Eval  & \multicolumn{3}{c}{Metrics}  \\
		\midrule
		\multicolumn{5}{c}{English Summarization} \\ \hline
		XSum & test & 47.7 & 24.6 & 39.7 \\
		CNNDM &      & 44.9 & 22.2 & 41.8 \\
		NewsRoom &      & 45.5 & 33.3 & 41.5 \\
		WikiHow &      & 46.4 & 22.1 & 45.2 \\ 
		SAMSum                        & & 54.6 & 30.3 & 46.1 \\
		Reddit TIFU                        &      & 31.0 & 11.6 & 25.3 \\ 
		AESLC                        &  & 38.9 & 22.5 & 37.7 \\
		MediaSum                        &  & 36.9 & 20.2 & 33.5 \\
		MultiNews                        &  & 47.9 & 36.8 & 43.9 \\
		arXiv                        &  & 50.0 & 22.5 & 44.9 \\
		PubMed                        &  & 51.1 & 24.9 & 46.9 \\ 
        \midrule
		\multicolumn{5}{c}{Multi-Lingual Summarization} \\ 
  \midrule
		WikiLingua(ru $\rightarrow$ en) & test & 38.8 & 15.9 & 32.7 \\
		WikiLingua(vi $\rightarrow$ en) &      & 39.3 & 16.7 & 33.2 \\
		WikiLingua(es $\rightarrow$ en) &      & 41.5 & 17.7 & 34.5 \\
		WikiLingua(tr $\rightarrow$ en) &      & 46.5 & 22.9 & 40.2 \\ 
  \midrule
		MLSum(de)                        & test & 47.9 & 36.8 & 43.9 \\
		MLSum(es)                        &      & 32.9 & 14.8 & 26.5 \\ 
		\bottomrule
	\end{tabular}
	   }
	\caption{
		ROUGE-1/ROUGE-2/ROUGE-L results on summarization tasks. 
	}
	\label{tab:rg}
	\vspace{-2mm}
\end{table}

\begin{table*}[htb!]
	\centering
	\begin{adjustbox}{max width=0.99\textwidth}
		\renewcommand{\arraystretch}{1.05}
		\begin{tabular}{l|cccc}
			\toprule
			{\bf Model}                & {0}                & {10}               & {100}    & {1000}  \\
			\midrule
			&
			\multicolumn{4}{c}{XSUM} \\ \midrule
			\modelp{T5}{LARGE}         & 12.8/2.3/9.8                & 13.2/2.5/10.0               & 21.5/5.5/17.0      & 31.2/9.4/23.8               \\
			\modelp{PEGASUS}{LARGE}    & 19.3/3.0/12.7                & 19.4/3.5/14.02                & 39.07/16.4/31.3     & 41.6/18.2/33.3                   \\
			\zcodepp$_{\mathtt{LARGE}}^{\dag}$            &    3.6/0.1/3.7            & 16.7/2.1/12.6                & 35.3/12.3/27.5     & 40.9/17.3/32.8                    \\ 
			\zcodepplarge{} & 36.6/13.7/28.6            & 37.4/14.0/29.1            & 40.6/17.5/30.0  & 41.9/18.9/33.6             
			\\ \midrule
			&
			\multicolumn{4}{c}{CNNDM} \\ \midrule
			\modelp{T5}{LARGE}         & 18.5/4.9/13.3                & 19.0/5.1/13.6                & 24.2/7.7/17.5      & 31.9/11.2/21.4                    \\
			\modelp{PEGASUS}{LARGE}    & 32.9/13.3/29.4               & 37.6/15.8/33.5               & 40.3/18.2/37.0     & 41.7/19.4/38.3                 \\
			\zcodepp$_{\mathtt{LARGE}}^{\dag}$            &    3.5/0.1/3.1            & 11.9/1.5/8.7                & 37.3/15.0/25.5     & 40.7/18.3/28.3                    \\
			\zcodepplarge{} & 40.0/17.3/25.3\rlap{$^*$} & 40.0/17.3/25.3\rlap{$^*$} & 41.1/18.4/27.5  & 42.0/19.6/28.9             \\
			\midrule
			&
			\multicolumn{4}{c}{SAMSum} \\ \midrule
			\modelp{T5}{LARGE}         & 9.4/1.3/8.2                & 14.0/4.0/12.0                & 29.6/10.4/23.5     & 41.4/17.8/32.8                   \\
			\modelp{PEGASUS}{LARGE}    & 26.3/6.4/20.5                & 37.0/11.7/28.1               & 45.0/19.8/36.1     & 49.3/24.4/40.6                \\
			\zcodepp$_{\mathtt{LARGE}}^{\dag}$            & 6.0/0.1/5.4             & 13.6/2.6/11.0                & 44.7/20.2/36.7     & 50.9/26.3/42.3                  \\
			\zcodepplarge{} & 26.5/7.9/20.5             & 40.27/17.4/33.7            & 47.6/22.3/38.7 & 52.2/28.1/43.9             \\
			\bottomrule
		\end{tabular}
	\end{adjustbox}
	\caption{ROUGE-1/ROUGE-2/ROUGE-L scores in different summarization datasets. Results are shown on their full test sets using 10, 100, and 1000 training examples. 0 denotes zero-shot results. Results marked with $^*$ mean that unfine-tuned checkpoints perform the best, i.e., zero-shot performance is better than the fine-tuned one. \zcodepp$_{\mathtt{LARGE}}^{\dag}$ refers to fine-tuning from phase 1 pre-trained model. \zcodepplarge{} fine-tuned from two-phase pre-trained model.}
	\label{tab:fewshot_evaluation}
	\vspace{-2mm}
\end{table*}

% \subsection{Rouge Scores on low-resource summarization tasks}

%\clearpage

\subsection{Fusion-in-Encoder structure}

\begin{figure}[htb!]

\centering  
\includegraphics[width=0.8\linewidth, trim={7.5cm 5.0cm 9.5cm 4cm},clip]{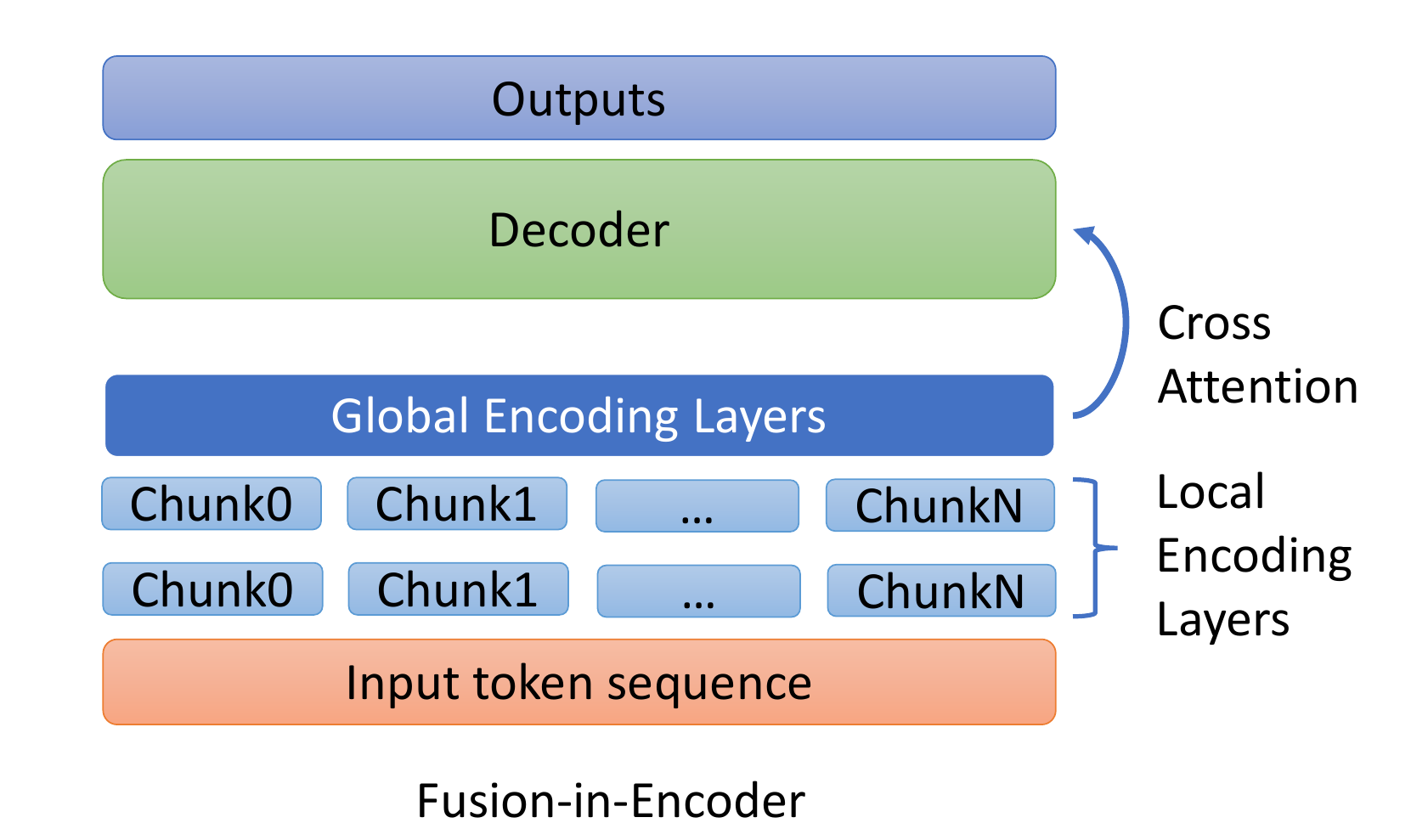}
\caption{The structure of Fusion-in-Encoder.}
\label{fig:fie}
\end{figure}

In figure \ref{fig:fie}, we show the architecture of FiE.

\label{app:fie}

\subsection{Ablation study}

We conducted a comprehensive experiment to explore what is important for the encoder's language understanding ability. Specifically, we experiment on the natural language inference task, e.g., MNLI \citep{mnli2018}, the question answering task, e.g., SQuAD \citep{squad1}, the summarization tasks, e.g., XSum \citep{narayan2018don} and CNNDM \citep{see2017get}. The results in Table \ref{tab:abs} show that using disentangled attention improves MNLI-matched/mismatched accuracy by 0.9\%/1.2\%, indicating an improvement in the encoder's language understanding ability. This improvement is also reflected in the performance of two summarization tasks, which see an improvement in R2 scores by 0.39\% and 0.22\%. Removing RTD significantly decreased performance, indicating that it is essential for improving the model's NLU capability. 
\begin{table*}[htb!]
	\centering
	\begin{adjustbox}{max width=0.95\textwidth}
		\begin{tabular}{@{\hskip1pt}l| c| c| c|c|c}
			\toprule
			\multirow{2}{*}{\bf Model}                    & {\#Traing} & {MNLI-m/mm}        & {SQuAD v1.1}       & {XSum}                     & CNNDM                 \\ 
			                                              & Tokens     & { Acc}             & F1/EM              & R1/R2/RL                   & R1/R2/RL        \\
			\midrule
			\modelp{T5}{BASE}                             & 1T         & 87.1/86.2          & 92.1/85.4          & 42.96/20.38/35.10 & 42.05/20.34/39.40                \\
   	\midrule
		\multicolumn{6}{c}{\emph{Our Implementations}} \\ 
		\midrule
  	ZCode++\textsubscript{BASE}                    & 130B       & \textbf{89.6/89.1} & \textbf{92.4/85.6} & \textbf{44.04/21.05/36.00} & \textbf{43.45/20.71/40.31} \\
   - DA                                          &            & 88.4/88.2          & 91.5/84.4          & 43.58/20.66/35.83          & 43.24/20.49/40.09          \\
   - DA - RTD         &            & 87.3/86.9          & 90.5/83.5          & 43.31/20.28/35.32          & 43.10/20.35/39.93          \\

  % + RTD                                          &            & 88.4/88.2          & 91.5/84.4          & 43.58/20.66/42.26          & 43.24/20.49/40.09          \\
  %   + RTD + DA		                    & 130B       & \textbf{89.6/89.1} & \textbf{92.4/85.6} & \textbf{44.04/21.05/36.00} & \textbf{43.45/20.71/40.31} \\			

			\bottomrule
		\end{tabular}
	\end{adjustbox}
	\caption{
		Ablation study of the impact of encoder performance on generation tasks.
	}
	\label{tab:abs}
	%\vspace{-3mm}
	  
\end{table*}

\subsection{Evaluate on NLU tasks}
In order to assess the model's effectiveness on natural language understanding (NLU) tasks, we conducted experiments using the eight NLU tasks from the GLUE dataset~\citep{wang2018glue}. These tasks are commonly used to evaluate sentence classification performance in machine learning. Our model, {\ModelName}, was tested using two approaches: adapting only the encoder and fine-tuning with a classification head, similar to BERT, or adapting the encoder-decoder and treating the task as a generation task, similar to T5. We compared {\ModelName} to other encoder-based PLMs with similar structures, including BERT, RoBERTa, ELECTRA, DeBERTa, and DeBERTaV3, as well as T5 for the encoder-decoder comparison. 

The results, shown in Table~\ref{tab:glue}, demonstrate that {\ModelName} performs comparably or better than the other models on all tasks. In particular, {\ModelName} outperformed the other encoder PLMs by an average of more than 1\% and outperformed T5 on all tasks with an average improvement of 1.98\% in test scores. These results demonstrate {\ModelName} as a strong universal language model with excellent performance on generation tasks and superior performance on NLU tasks.
\begin{table*}[htb!]
	\centering
	\begin{tabular}{@{\hskip3pt}l@{\hskip2pt}|@{\hskip2pt} c@{\hskip2pt}|@{\hskip2pt} c@{\hskip2pt}| @{\hskip2pt}c@{\hskip2pt}|c@{\hskip2pt}|c@{\hskip2pt}|c@{\hskip2pt}|c@{\hskip2pt}|c@{\hskip2pt}|c@{\hskip2pt}|c@{\hskip2pt}}
		\toprule
		\multirow{2}{*}{\bf Model} & {Eval} & {CoLA}   & {QQP}    & {MNLI-m/mm}        & SST-2    & STS-B    & QNLI          & RTE           & MRPC     & Avg.                    \\ 
		& & Mcc & Acc & Acc & Acc &Corr&Acc&Acc&Acc \\
		\#Train    && 8.5k & 364k & 393k & 67k &7k&108k&2.5k&3.7k \\
		\midrule
		\multicolumn{11}{c}{\emph{Encoder-Only}} \\ 
		\midrule
		\modelp{BERT}{LARGE}       & Dev    & 60.6     & 91.3     & 86.6/-             & 93.2     & 90.0     & 92.3          & 70.4          & 88.0     & 84.05                   \\ 
		\modelp{RoBERTa}{LARGE}    &        & 68.0     & 92.2     & 90.2/90.2          & 96.4     & 92.4     & 93.9          & 86.6          & 90.9     & 88.82                   \\ 
		% T5\textsubscript{large-test} &52.1 & 89.9  & 90.2/89.6 & 96.2 &86.6&95.6&75.7&87.2&  \\ 
		\modelp{ELECTRA}{LARGE}    &        & 69.1     & 92.4     & 90.9/-             & 96.9     & 92.6     & 95.0          & 88.0          & 90.8     & 89.46                   \\ 
		\modelp{DeBERTa}{LARGE}    &        & 70.5     & 92.3     & 91.1/91.1          & 96.8     & 92.8     & 95.3          & 88.3          & 91.9     & 90.00                   \\ 
		\modelp{DeBERTaV3}{LARGE}  &        & 75.3     & \bf 93.0 & \textbf{91.8/91.9} & \bf 96.9 & 93.0     & \textbf{96.0} & \textbf{92.7} & 92.2     & \textbf{91.37}          \\ 
		\ModelName{}         &        & \bf75.5  & 92.8     & 91.7/91.5          & 96.3     & \bf93.1  & 95.8          & 92.4          & \bf92.4  & 91.23                   \\ 
		\midrule
		\multicolumn{11}{c}{\emph{Encoder-Decoder}} \\ 
		\midrule
		\modelp{T5}{LARGE}         & Test   & 61.2     & 89.9     & 89.9/89.6          & 96.3     & 89.9     & 94.8          & 87.2          & \bf 89.9 & 87.35                   \\ 
		\ModelName{}         & Test   & \bf 69.2 & \bf 90.0 & \bf 91.0/90.9      & \bf 97.9 & \bf 91.2 & \bf 95.1      & \bf 90.7      & 89.6     & \bf89.33                \\
		\ModelName{}         & Dev    & 86.2     & 92.4     & 91.4/91.4          & 96.5     & 92.5     & 95.2          & 92.1          & 91.2     & 92.19                   \\
		\bottomrule
	\end{tabular}
	\caption{
		Comparison results on the GLUE development set. To make a fair comparison, following previous work on encoder models, we evaluate {\ModelName} with development set. For Encoder-Decoder model we follow T5 to fine-tune all tasks jointly and submit result on test set to GLUE evaluation server.  
	}
	\label{tab:glue}
	%\vspace{-2mm}
\end{table*}

\subsection{Evaluate on NLG tasks}

We evaluated the language generation performance of \modelp{\ModelName}{} on a range of English tasks, including abstractive document summarization tasks (XSum, CNNDM, Wikilingual-en), a conversational summarization task (SAMSum), data-to-text tasks (WebNLG-en, E2ENLG) and a question answering task (SQuAD v1.1). We compared the performance of the \modelp{\ModelName}{} model to other state-of-the-art models with similar architectures and parameters, as shown in Table \ref{tab:en_nlg}.

\newcolumntype{g}{>{\columncolor{Gray}}c}
\definecolor{Gray}{gray}{0.97}
\begin{table*}[htb!]
	\centering
	\begin{adjustbox}{max width=0.95\textwidth}
		\renewcommand{\arraystretch}{1.05}
		\begin{tabular}{l|c|cccc|ccc|g}
			\toprule
									
			\multirow{2}{*}{\bf Dataset} & \multirow{2}{*}{Metric} & \modelp{BART}{LARGE} & \modelp{PEGASUS}{LARGE} & \modelp{T5}{LARGE} & \modelp{T5}{XLARGE} & \modelp{PaLM}{} & \modelp{GPT3}{} & \modelp{UL2}{}    & \ModelName{}      \\ 
			                             &                         & 400M                 & 500M                    & 800M               & 3B                  & 540B            & 175B            & 20B               & 800M                    \\
			\midrule
			XSum                         & R1/R2/RL                & 45.1/22.3/37.3       & 47.2/24.6/39.4          & 44.3/22.0/36.7     & -                   & -/21.2/-        & -               & \textbf{-/26.6/-} & \textbf{47.7/24.7/39.7} \\
			CNNDM                        & R1/R2/RL                & 44.2/21.3/40.9       & 44.2/21.5/41.1          & 43.6/21.4/40.6     & 42.7/21.0/39.9      & -               & -               & -/21.9/-          & \textbf{44.9/22.0/41.8} \\
			SAMSum                       & R1/R2/RL                & 53.4/28.7/44.2       & 50.2/26.3/46.2          & 51.0/27.0/\bf 46.6 & -                   & -               & 53.8/29.8/45.9  & -/29.6/-          & \textbf{54.6/30.3}/46.1 \\
			\midrule
			WebNLG-en                    & R1/R2/RL                & -                    & -                       & 67.1/39.6/51.8     & 75.4/49.4/59.5      & -/49.3/-        & -               & -/55.4/-          & \textbf{79.0/56.3/64.6} \\
			E2E NLG                      & R1/R2/RL                & -                    & -                       & 70.8/41.7/49.5     & 70.8/41.7/49.7      & -/45.3/-        & -               & -/46.5/-          & \textbf{74.8/46.9/54.0} \\
			
			\bottomrule
		\end{tabular}
	\end{adjustbox}
 \caption{Comparison results on English NLG tasks. }
				 
	% \caption{Results on Common English Summarization tasks. Best numbers are in \textbf{Bold}. \textsuperscript{a}ST-MOE\textsubscript{268B} \citep{zoph2022designing}, \textsuperscript{b}T5\textsubscript{11B} \citep{rothe2021thorough}, \textsuperscript{c}GPT3\textsubscript{175B+LoRA} \citep{hu2021lora},
	% \textsuperscript{d}MAPPET+BART\textsubscript{LARGE} \citep{aghajanyan2021muppet} }
					
	\label{tab:en_nlg}
	% \vspace{-2mm}
\end{table*}

Results show that \ModelName{} outperforms all of the other models' scores by a large margin in terms of ROUGE and BLEU scores. For example, \ModelName{} significantly outperformed \modelp{T5}{XLARGE} on CNNDM by 1\% in terms of ROUGE-2 score, on the WebNLG-en task by 6.9\%, and about 1\% BLEU score on dialog response generation tasks. Even though it has less than 1/3 the parameters of \modelp{T5}{XLARGE}, \ModelName{} outperformed PEGASUS on SAMSum task by 4\% in terms of ROUGE-2 score. We conjecture that PEGASUS is a model specifically optimized for summarization using 1500GB of news data, which may have introduced a domain mismatch with the conversational summarization task. We also compared \ModelName{} to other state-of-the-art models with extremely large parameters, including PaLM, GPT3, and UL2. \ModelName{} outperformed PaLM on three out of four tasks by a large margin, even though it has less than 1/600 the parameters of PaLM. \ModelName{} also outperformed \modelp{UL2}{20B} on four out of five tasks, even though it has less than 1/20 the parameters of \modelp{UL2}{20B}. These results demonstrate the efficiency of the {\ModelName} model.

\end{document}